\documentclass[10pt]{article} 
\usepackage{multirow}
\usepackage{amsmath}
\usepackage[linesnumbered,ruled,vlined]{algorithm2e}
\usepackage{float}
\usepackage{enumitem}
\usepackage{pifont}
\newcommand{\cmark}{\ding{51}}%
\newcommand{\xmark}{\ding{55}}%
\usepackage{color, colortbl}
\usepackage{xcolor}

\newcommand{\methodname}[0]{SAP}
\newcommand{\abs}[1]{\lvert#1\rvert}

\definecolor{Gray}{gray}{0.8}
\usepackage{xcolor}
\definecolor{darkgreen}{rgb}{0.0, 0.5, 0.0}
\definecolor{darkred}{rgb}{0.82, 0.1, 0.26}
\definecolor{classcolor}{HTML}{66B2FF}
\definecolor{attrcolor}{HTML}{FF6666}
\usepackage{lipsum} 
\usepackage{subcaption}
\usepackage{wrapfig}
\usepackage{graphicx}
\usepackage{hyperref}       
\hypersetup{
    colorlinks=true,
    linkcolor=red,
    citecolor=blue,
    filecolor=magenta,      
    urlcolor=cyan,
    }
\usepackage{booktabs}

\usepackage[export]{adjustbox}
\usepackage[accsupp]{axessibility}  
\usepackage{ulem}

\usepackage{hyperref}

\usepackage[preprint]{tmlr}

\usepackage{amsmath,amsfonts,bm}

\def\eqref#1{equation~\ref{#1}}

\def\1{\bm{1}}

\DeclareMathAlphabet{\mathsfit}{\encodingdefault}{\sfdefault}{m}{sl}
\SetMathAlphabet{\mathsfit}{bold}{\encodingdefault}{\sfdefault}{bx}{n}

\DeclareMathOperator*{\argmax}{arg\,max}

\usepackage{hyperref}
\usepackage{url}

\title{Semantic Alignment for Prompt-Tuning in Vision Language Models}

\author{\name Hari Chandana Kuchibhotla* \email ai20resch11006@iith.ac.in \\
 \addr Indian Institute of Technology Hyderabad, India
      \AND
      \name Sai Srinivas Kancheti* \email cs21resch01006@iith.ac.in \\
       \addr Indian Institute of Technology Hyderabad, India
      \AND
      \name Abbavaram Gowtham Reddy \email cs19resch11002@iith.ac.in\\
       \addr Indian Institute of Technology Hyderabad, India
      \AND
      \name Vineeth N Balasubramanian \email vineethnb@iith.ac.in \\
      \addr Indian Institute of Technology Hyderabad, India}

\def\month{MM}  
\def\year{YYYY} 
\def\openreview{\url{https://openreview.net/forum?id=XXXX}} 

\begin{document}

\maketitle

\begin{abstract}
Going beyond mere fine-tuning of vision-language models (VLMs), learnable prompt tuning has emerged as a promising, resource-efficient alternative. Despite their potential, effectively learning prompts faces the following challenges: (i) training in a low-shot scenario results in overfitting, limiting adaptability, and yielding weaker performance on newer classes or datasets; (ii) prompt-tuning's efficacy heavily relies on the label space, with decreased performance in large class spaces, signaling potential gaps in bridging image and class concepts. In this work, we investigate whether better text semantics can help address these concerns. In particular, we introduce a prompt-tuning method that leverages class descriptions obtained from Large Language Models (LLMs). These class descriptions are used to bridge image and text modalities. Our approach constructs part-level description-guided image and text features, which are subsequently aligned to learn more generalizable prompts. Our comprehensive experiments conducted across 11 benchmark datasets show that our method outperforms established methods, demonstrating substantial improvements.
\end{abstract}

\section{Introduction}
\def\thefootnote{*}\footnotetext{Equal contribution}\def\thefootnote{\arabic{footnote}}

Foundational Vision-Language Models (VLMs) like CLIP~\citep{clip} and ALIGN~\citep{align} have displayed remarkable zero-shot and open-vocabulary capabilities in recent years. This has led to VLMs being employed in various vision-only downstream tasks such as open-vocabulary image classification~\citep{openvocab}, object detection~\citep{feng2022promptdet}, and image segmentation~\citep{segmentation}. Trained on extensive web data, these models often use a contrastive loss to align image-text pairs in a shared embedding space, allowing them to represent diverse concepts. 

Recently, learnable prompt-tuning~\citep{coop,promptsrc,poda} has emerged as a promising parameter-efficient alternative for fine-tuning foundation models. Prompt-tuning methods introduce additional learnable parameters called \textit{prompt vectors}, which are tuned on task-specific data. This approach adapts VLMs for a specific downstream task without affecting the pre-trained parameters of the VLM. While prompt-tuning methods have shown great promise, efficiently learning prompt vectors faces the following challenges: (i) training prompts in a low-shot setting leads to overfitting, hindering their generalizability, and exhibiting sub-optimal performance when applied to newer classes or datasets~\citep{logoprompt, promptsrc, maple}, (ii) the performance of prompt-tuning methods can be highly dependent on the label space used for classification. During inference, if the label space is large, the performance tends to decrease due to bias towards the seen classes the model was fine-tuned on (see empirical evidence in Tab.~\ref{gzs_table} of \S~\ref{sec experiments}). These issues indicate that there is a lack of understanding of images and classes based on their detailed semantic components. For example, an image of a cat should be understood through its specific features like `whiskers' and `tail', not just the class name `cat.'
To address this, we propose \textit{SAP} (\textbf{S}emantic \textbf{A}lignment for \textbf{P}rompt-tuning), which uses class descriptions to learn generalizable prompts. Semantic alignment involves matching meaningful part-level image features with their corresponding text features. This alignment helps the model grasp the relationship between different parts of an image and their textual descriptions, leading to a more detailed and accurate representation.

\begin{figure}[t]
\centering
\begin{subfigure}{.99\textwidth}
  \centering
\includegraphics[width=0.95\linewidth]{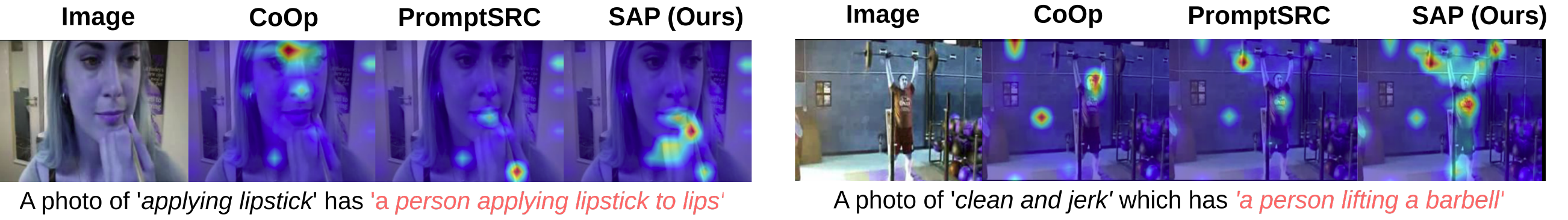}
\vspace{5pt}
\end{subfigure}
\\
\begin{subfigure}{.99\textwidth}
  \centering
\includegraphics[width=.95\linewidth]{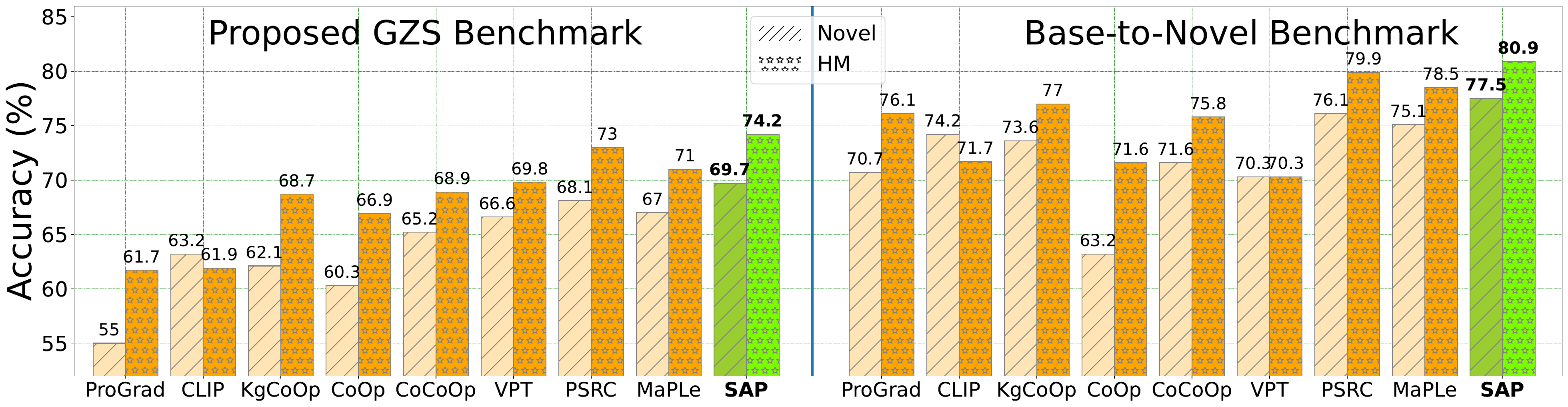}
\end{subfigure}
\vspace{-5pt}
\caption{\textbf{Top:} Comparison of GradCAM~\citep{gradcam} visualizations for our proposed method SAP against other baselines, on classes \textit{``Applying Lipstick''} and \textit{``Clean and Jerk''} from an Action Recognition dataset~\citep{ucf101}. The saliency maps indicate image regions that are most relevant to the
descriptions ``A photo of applying lipstick has a person applying lipstick to lips'' and ``A photo of clean and jerk which has a person lifting a barbell'' respectively. SAP effectively localizes the text semantics in images compared to baselines.
\textbf{Bottom:} SAP surpasses other baselines on Generalized Zero-Shot (GZS) and Base-to-Novel (B2N) benchmarks, showing improvements of \textbf{+1.6\%} and \textbf{+1.2} on Novel Accuracy and Harmonic Mean (HM) for GZS, and \textbf{+1.4\%} and \textbf{+0.9} for B2N compared to best performing baselines.}
\vspace{-10pt}
\label{fig:teaser}
\end{figure}

Our method uses class descriptions to guide the creation of such image and text features that correspond to specific parts or aspects of a class. However, we observe that merely using class descriptions alone does not address the challenges presented above, as shown in Tab. \ref{tab:text_analysis} of \S~\ref{sec:additional_results}. We demonstrate that careful \textit{semantic alignment} between image and text features is crucial for effectively leveraging class descriptions.
Given a set of class descriptions, we show how to construct \textit{description-guided} image and text features. For instance, for an image of a cat, and a class description `has a large tail', the corresponding description-guided image feature encodes the part-level visual semantic information related to the description. We then compute semantic alignment as the average cosine similarity between description-guided image and text features, for relevant descriptions. We use pre-trained Large Language Models (LLMs) to generate class descriptions in an inexpensive manner. A recent set of works~\citep{iclr23,labo} has shown that class descriptions obtained from LLMs can be naively used to classify images on a given dataset with fixed categories. We go beyond and leverage these class descriptions to perform low-resource prompt-tuning, and show that such adapted VLMs show better generalization to unseen, novel classes. Fig. \ref{fig:teaser} illustrates the effectiveness of SAP over other baselines on two benchmarks, Generalized Zero-Shot Classification (GZS) and Base-to-Novel Classification (B2N), defined in \S~\ref{sec experiments}. As our semantic alignment is part-level, SAP also showcases superior localization of visual concepts relevant to a class description, as seen through class activation maps, when compared to other baselines.

Tab. \ref{tab:comparison} delineates the key differentiators of our approach compared to other baselines. Most existing prompt-tuning methods do not use additional text semantics; even among the recent few that use such information, our method utilizes class descriptions at a part-level for both image and text. 
This strategy leads to non-trivial performance enhancements across benchmark datasets and improved localizations in novel classes or datasets. Additionally, we highlight a gap in the evaluation scheme used in existing prompt-tuning efforts, which 
demonstrate the performance of learned prompts primarily on the tasks of Base-to-Novel classification and cross-dataset evaluation. Inspired by the traditional Generalized Zero-Shot Learning (G-ZSL)~\citep{traditionalzsl,zslcvpr23} paradigm, we posit that generalization in the zero-shot setting is more realistic when considering both base and novel classes at inference. We call this protocol \textit{GZS evaluation} -- the first such effort among prompt-tuning methods. We also propose another benchmark --  \textit{Out-of-Vocabulary Classification} -- where the method is exclusively evaluated using class descriptions to classify images when its label lies outside CLIP's vocabulary. Our contributions can be summarized as follows.

\begin{itemize}[leftmargin=*]
\item We propose a prompt-tuning method to fine-tune VLMs that can leverage class descriptions obtained from an LLM. Our novel approach to combine class descriptions with visual part-level information allows us to utilize local features of an image, thus bridging image and text modalities using class descriptions. This improved alignment allows us to learn prompts that can generalize well to unseen classes and datasets.
\item We carry out a comprehensive suite of experiments with comparisons against state-of-the-art and very recent methods on eleven standard benchmark datasets. We outperform existing baselines with a significant margin on all evaluation protocols.
\item We propose two new evaluation protocols: GZS evaluation and Out-of-Vocabulary Classification to better study the generalizability of prompt-tuning methods for VLMs. Our method consistently outperforms earlier baselines on these protocols, too.
\end{itemize}
\begin{table}
    \centering
    \scalebox{0.55}{
    \begin{tabular}{l|cccccc}
    \toprule
    &\textbf{Text}&\textbf{Image}&\textbf{Use of External}&\textbf{Part-level }&\textbf{Evaluation}&\textbf{\# of Additonal }\\
    \textbf{Method}&\textbf{Prompts}&\textbf{Prompts}&\textbf{Knowledge}&\textbf{img-text}&\textbf{Benchmarks}&\textbf{Trainable}\\
    &&&&\textbf{alignment}&&\textbf{Parameters}\\
    \midrule
     \textbf{CoOp} [IJCV '22], \textbf{CoCoOp} [CVPR '22], \textbf{KgCoOp} [CVPR '23]& \Large \textcolor{blue}{\cmark} & \Large \textcolor{red}{\xmark}&\Large \textcolor{red}{\xmark}&\Large \textcolor{red}{\xmark}&B2N, XDataset, DG& 2k - 36k\\ 
     \textbf{ProGrad} [ICCV '23], \textbf{ProDA} [CVPR '22]&&&&& \\
     \midrule
     \textbf{MaPLe} [CVPR '23], \textbf{PSRC} [ICCV '23], \textbf{LoGoPrompt} [ICCV '23]&\Large \textcolor{blue}{\cmark} &\Large \textcolor{blue}{\cmark}&\Large \textcolor{red}{\xmark}&\Large \textcolor{red}{\xmark}&B2N, XDataset, DG& 36k - 3.55M\\ 
     \midrule
     \textbf{KAPT} [ICCV '23], \textbf{CoPrompt} [ICLR '24], \textbf{CLIP-VDT} [ICCVW '23]&\Large \textcolor{blue}{\cmark}&\Large \textcolor{blue}{\cmark}&\Large \textcolor{blue}{\cmark}&\Large \textcolor{red}{\xmark}&B2N, XDataset, DG&1.3M - 4.74M\\
     \midrule
      \rowcolor{teal!15}
     \textbf{SAP (Ours)} &\Large \textcolor{blue}{\cmark}&\Large \textcolor{blue}{\cmark}&\Large \textcolor{blue}{\cmark}&\Large \textcolor{blue}{\cmark}& GZS, B2N, XDataset, DG,&\textbf{36K}\\
     \rowcolor{teal!15}
     &&&&&Out-of-Vocabulary Classification&\\
    \bottomrule
    \end{tabular}
    }
    \vspace{5pt}
    \caption{Comparison of the proposed method, SAP, with other related work on various key aspects involving fine-tuning VLMs for better generalization. B2N: Base-to-Novel, XDataset: Cross Dataset, DG: Domain Generalization, GZS: Generalized Zero-Shot.}
    \label{tab:comparison}
\end{table}

\section{Related Work}

\textbf{Vision-Language Models.} Vision-language models (VLM) exhibit significant promise in acquiring generic visual representations. VLMs aim to harness natural language guidance for image representation learning and concurrently align both the text and image features within a shared embedding space. We consider encoder-only VLMs which comprise of three components: a text encoder, an image encoder, and a learning methodology that effectively utilizes information from both text and image modalities. Recent research on learning transferable visual representations delves into establishing semantic connections between text and visual elements, capitalizing on a vast reservoir of internet-based image-text pairs. For instance, CLIP~\citep{clip} is the product of contrastive learning from 400 million image-text pairs, while ALIGN~\citep{align} utilizes 1.8 billion noisy image-text pairs extracted from raw alt-text data. Nonetheless, a substantial challenge persists in transferring these foundational models to downstream tasks while preserving their initial capacity for generalization. To address this, we use auxiliary information in the form of class descriptions to better align image and text features, thereby enhancing the model's performance and generalizability.

\noindent\textbf{Prompt-Tuning.}
Prompt-tuning introduces task-specific text tokens designed to be learnable to customize the pre-trained VLM for downstream tasks. Context Optimization (CoOp)~\citep{coop} marks the pioneering effort in replacing manually crafted prompts with adaptable soft prompts, fine-tuned on labeled few-shot samples. Conditional Context Optimization (CoCoOp)~\citep{cocoop} builds upon this by generating image-specific contexts for each image and merging them with text-specific contexts for prompt-tuning. In contrast, Visual Prompt Tuning \citep{vpt} introduces learnable prompts exclusively at the vision branch, resulting in sub-optimal performance for transferable downstream tasks. ProDA~\citep{proda} focuses on learning the distribution of diverse prompts. KgCoOp~\citep{kgcoop} introduces regularization to reduce the discrepancy between learnable and handcrafted prompts, enhancing the generalizability of learned prompts to unseen classes. PSRC~\citep{promptsrc} shares a similar concept with KgCoOp~\citep{kgcoop} but introduces Gaussian prompt aggregation. ProGrad~\citep{prograd} selectively modifies prompts based on gradient alignment with a hard-coded prompt. MaPLe~\citep{maple} introduces prompts at text and image encoder branches and link them with a coupling function. In a different approach, LoGoPrompt~\citep{logoprompt} capitalizes on synthetic text images as effective visual prompts, reformulating the classification problem into a min-max formulation. Although these methods have shown promising results, they suffer from overfitting to the training classes when trained in a low-shot manner. This overfitting limits their generalizability and results in sub-optimal performance on newer classes or datasets. We address this issue by leveraging external information in the form of class descriptions to semantically align image and text features,  helping us learn generalizable prompts.

\noindent\textbf{Use of External Knowledge.} A set of recent works~\citep{iclr23,labo,Pratt2022WhatDA} provide evidence that visual recognition can be improved using concepts, and not just class names. However,~\citep{iclr23,Pratt2022WhatDA} does not facilitate a way to perform fine-tuning on a downstream dataset. In contrast,~\citep{labo} is a concept bottleneck model with a fixed label space and thus cannot be used for zero-shot classification. In fine-tuning methods incorporating external knowledge, KAPT~\citep{kapt} introduces complementary prompts to simultaneously capture category and context but lacks semantic alignment of each class description at the part-level of both image and text. On the other hand, CLIP-VDT~\citep{vdt_adapter} utilizes semantic-rich class descriptions only in the text modality, without semantic alignment with images. In CoPrompt\citep{coprompt}, class descriptions are utilized via a regularizer acting as a consistency constraint to train the text prompts. There is no consideration of explicit semantic alignment with the image modality. In contrast to existing methods, our approach utilizes class descriptions to semantically construct both text and image features, enhancing part-level alignment between the two modalities. This improved alignment helps us learn prompts that can generalize well to unseen classes and datasets. A comparison of our method with existing works is shown in Tab.~\ref{tab:comparison}.

\section{Preliminaries and Background}
\label{sec:prelims}

VLMs perform image classification on a downstream dataset by comparing an image representation with text representations of the class names in the dataset's label space. When a small amount of labeled data is available, it has been shown that fine-tuning VLMs substantially boosts downstream performance~\citep{coop,cocoop}. However, the fine-tuned model does not generalize to novel classes that were absent during fine-tuning~\citep{cocoop}. In this work, we propose \textbf{S}emantic \textbf{A}lignment for \textbf{P}rompt Learning  (SAP), that leverages class descriptions to fine-tune VLMs for better generalization to novel classes. Before we describe our methodology, we briefly discuss the required preliminaries, beginning with CLIP~\citep{clip}, the VLM chosen as our backbone following earlier work~\citep{coop,cocoop,maple,proda,kgcoop,promptsrc,prograd}. A summary of notations and terminology is presented in Appendix \S~\ref{app summary of noations}.

\vspace{3pt}
\noindent \textbf{CLIP Preliminaries.} 
CLIP consists of an image encoder $\theta$ and a text encoder $\phi$, which are trained contrastively on paired image-text data to learn a common multi-modal representation space. $\theta$ takes an image $\mathbf{x}$ as input and returns the image feature $\theta(\mathbf{x})\in\mathbb{R}^d$. $\phi$ processes a text string $S$ into a $d$-dimensional feature vector $\phi(S)\in\mathbb{R}^d$. CLIP is trained with InfoNCE loss~\citep{Oord2018RepresentationLW} to enhance cosine similarity for matching image-text pairs and to reduce it for non-matching pairs.

CLIP performs zero-shot visual recognition of an image $\mathbf{x}$ by choosing the most similar class name from a set of candidate class names $\mathcal{Y}$, i.e., predicted class $\hat{y}=\argmax_{y\in\mathcal{Y}} sim(\theta({\mathbf{x}}), \phi(y))$, where the similarity measure $sim$ is cosine-similarity. In practice, for a class name $y$, $\phi(y)$ is the text representation of a manually crafted prompt encapsulating $y$ such as \texttt{`a photo of a [y]'}. Zero-shot classification performance significantly depends on the label set $\mathcal{Y}$ considered, and can also vary with the template of the text prompt~\citep{clip}.


\vspace{3pt}
\noindent \textbf{Fine-Tuning CLIP with Learnable Prompts.} 
To perform efficient adaptation under limited supervision, prompt-tuning methods add a small number of learnable tokens to the input token sequence of either modality which are fine-tuned to generate task-specific representations. For instance, CoOp~\citep{coop} adds $n$ learnable text-prompts $\rho_t=\{\mathbf{p}_1^{t},\dots,\mathbf{p}_n^{t}\}$ to the token embeddings $\{\mathbf{w}_1^S,\dots,\mathbf{w}_q^S\}$ of some text $S$. The final sequence $\{\mathbf{p}_{1}^{t},..,\mathbf{p}_{n}^{t},\mathbf{w}_{1}^S,..,\mathbf{w}_{q}^S\}$ is passed through $\phi$ to obtain the \textit{prompted text feature} $\phi_p(S)$\footnote{\scriptsize We add a subscript $p$ to indicate prompted features for images and text}.  
We follow IVLP~\citep{ivlp}, which adds learnable prompt tokens at transformer layers of both image and text encoders. That is, along with text prompts, IVLP appends learnable visual prompts $\rho_v$ to patch tokens of image $\mathbf{x}$, which are passed through $\theta$ to yield the \textit{prompted visual feature} $\theta_p(\mathbf{x})$. Let $\rho =\{\rho_t, \rho_v\}$ denote the set of all trainable text and visual prompts. These prompts are trained to maximize the similarity between a prompted image feature and the corresponding prompted text feature of its class label. Given $B$ image-text pairs $\{(\mathbf{x}_i,y_i)\}_{i=1}^B$, where $y_i\in \mathcal{Y}$, the likelihood of $\mathbf{x}_i$ predicting class $y_i$ is given by $\displaystyle \mathbb{P}_{\rho}(y_i\mid \mathbf{x}_i)= \frac{\exp(sim(\theta_p(\mathbf{x_i}), \phi_p(y_i))/\tau)}{\sum\limits_{y\in\mathcal{Y}} \exp( sim(\theta_p(\mathbf{x_i}), \phi_p(y))/\tau)}$, where $\tau$ is the temperature and $sim$ is cosine similarity. The negative log-likelihood loss to be optimized is $L(\rho) = \frac{-1}{B}\sum\limits_{i\in[B]} \log(\mathbb{P}_{\rho}(y_i\mid \mathbf{x}_i))$.

With the above background, we now present our methodology to use class descriptions to learn prompts that helps VLMs generalize better to unseen, novel classes.

\section{Semantic Alignment for Prompt-tuning: Methodology}

\setlength{\intextsep}{0pt}
\begin{figure}
    \centering
    \includegraphics[width=1.0\textwidth]{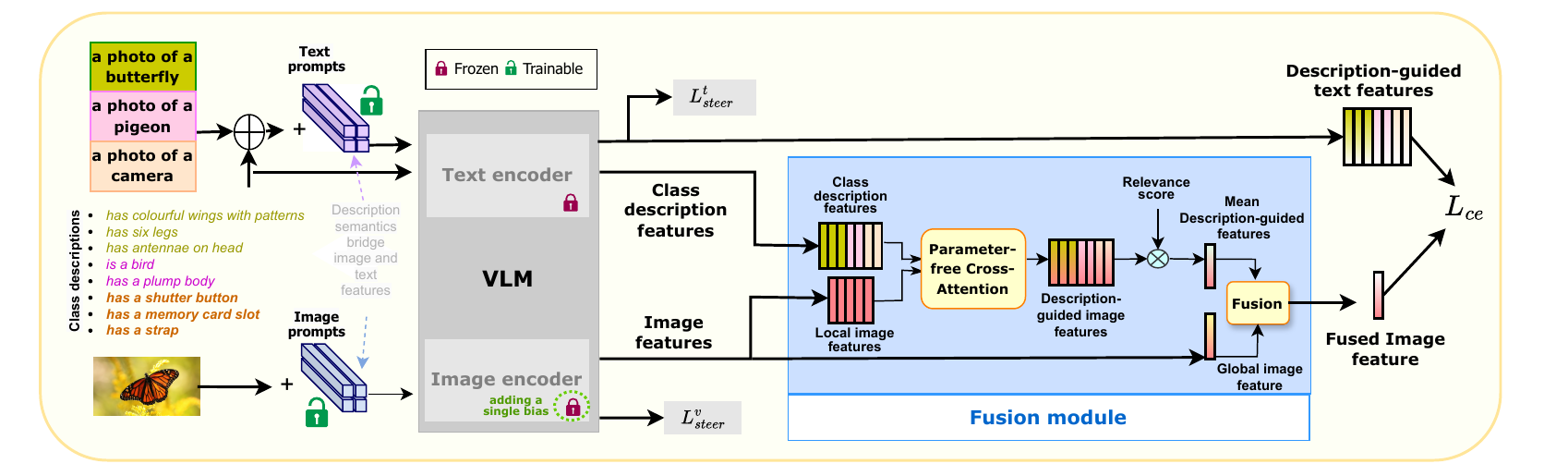}
    \caption{Our proposed workflow, SAP, performs part-based semantic alignment between image and text features. SAP integrates class descriptions into the text template which are passed through the text encoder to construct description-guided text features. Global and local image features are obtained from the image encoder. Description-guided image features are obtained by performing parameter free cross-attention between class descriptions and local features. These image features are pooled into a mean description-guided image feature, which is then fused with the global image feature to obtain the fused image feature. Description-guided text features and the fused image feature contain part-level semantic information, and are semantically aligned. We optimize a cross-entropy loss $L_{ce}$, and two steering losses $L_{steer}^{v}$, and $L_{steer}^{t}$.}
    \label{fig:mainfigure}
\end{figure}

Given labeled data, most existing methods learn prompts that largely limit themselves to incorporating text information in the form of class labels only. We propose SAP, \textbf{S}emantic \textbf{A}lignment for \textbf{P}rompt-tuning, which utilizes auxiliary information in the form of class descriptions obtained from LLMs to learn more generalizable prompts. Our method constructs description-guided image and text features that are semantically aligned with each other. Specifically, a class description provides a semantic context, and the corresponding description-guided image or text feature encodes part-level information related to this description. Semantic Alignment is thus the process of matching meaningful part-level image features with their corresponding description features.
This external semantic knowledge, derived from class descriptions, transfers to novel classes because the semantics represent common concepts shared across multiple classes, such as `large tail' or `whiskers'. The overview of our methodology is shown in Figure~\ref{fig:mainfigure}. We begin by describing how class descriptions are generated using LLMs.

\vspace{-6pt}
\subsection{Generating Class Descriptions}
\label{sec:generating_class_descriptions}
\vspace{-5pt}
Large language models (LLMs) act as vast knowledge corpora that can be queried for the semantics of real-world objects. We use the popular LLM GPT-3.5~\citep{gpt35} to obtain text descriptions for each class in a given dataset. 
Class descriptions commonly contain visual cues such as shape, texture, and color, as well as narratives of objects commonly correlated with the class. To keep our method cost-efficient, we use descriptions that are class-specific but not image-specific, thus making them reusable for a set of image samples (note that this is done only once per class label). We use the responses from the LLM as they are, and do not manually curate or filter them any further. This keeps our approach low-cost while integrating finer semantic details into fine-tuning of VLMs.
Some examples of our class descriptions are provided in Appendix \S~\ref{app sec attribute priors}.

\noindent \textbf{Class Description Features.} For each class $y\in \mathcal{Y}$, where $\mathcal{Y}$ is the label space under consideration, we denote by $A_y$ the set of generated class descriptions. Let $A = \bigcup\limits_{y\in\mathcal{Y}} A_y$,  $N=|A|$ denote the set of descriptions of all classes and the size of the set, respectively. \textit{Class description features} $\phi(A)\in\mathbb{R}^{N\times d}$ are obtained by passing the class descriptions through text-encoder $\phi$. In the following sections, we describe how SAP leverages class descriptions to construct description-guided image and text features, enabling us to learn prompts that generalize well.

\vspace{-7pt}
\subsection{Leveraging Class Descriptions for Text Features}
\label{sec:leveraging_descriptions_for_text_features}
\vspace{-6pt}
The text feature $\phi(y)$ for a class $y\in\mathcal{Y}$ is generally obtained by encapsulating the class name in a text template, for eg. \texttt{`a photo of a [y]'}, and passing it through $\phi$. When class descriptions $A_y$ are given, we append them to the text template to generate $\abs{A_y}$ distinct templates. For example for class $y=cat$ and $A_y=\{\text{`has whiskers'}, \text{`has a large tail'}\}$, we generate $2$ description-guided templates \texttt{`a photo of a cat which has whiskers'} and \texttt{`a photo of a cat which has a large tail'}.

The description-guided templates are passed through text-encoder $\phi$ to generate description-guided text features $\phi(y;A_y)\in\mathbb{R}^{|A_y|\times d}$ for class $y$. For an image $\mathbf{x}$, the semantic alignment $\xi$ between the image feature $\theta(\mathbf{x})$ and description-guided text features for class $y$ is given by:
\begin{equation}
\label{eq:semantic alignment}
    \xi(\theta(\mathbf{x}), \phi(y;A_y)) = \frac{1}{\abs{A_y}}\sum\limits_{a\in A_y} sim(\theta(\mathbf{x}), \phi(y;a))
\end{equation}
We find that this simple way of incorporating class descriptions into the text modality works well in practice. We validate our design choices in Tab.~\ref{tab:text_analysis} by comparing against alternative ways to incorporate class descriptions.

\vspace{-7pt}
\subsection{Leveraging Class Descriptions for Image Features}
\label{sec:leveraging_descriptions_for_image_features}

\begin{wrapfigure}[10]{r}{0.42\textwidth}
    \centering
    \includegraphics[width=0.5\textwidth]{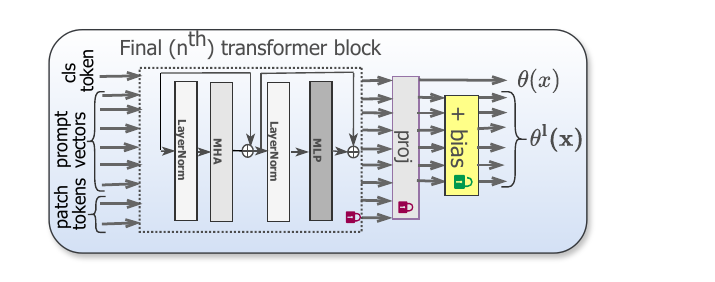}
    \caption{Addition of a bias vector to the last transformer block in $\theta$}
    \vspace{7pt}
    \label{bias_figure}
\end{wrapfigure}
\vspace{-6pt}
As shown in the fusion module of Fig.~\ref{fig:mainfigure}, we also leverage the class descriptions in the visual modality by first generating \textit{description-guided} image features, and then fusing them with the \textit{global} image feature. We describe this process below.

\vspace{2pt}
\noindent \textbf{Constructing Description-Guided Image Features.}

An image $\mathbf{x}$ is passed through the image encoder $\theta$ (which is a vision transformer in this section) and the output of the final transformer block of shape $(1+196+n)\times d^{\prime}$ is collected. Here, $1$ corresponds to the $\textbf{cls}_{\mathcal{I}}$ token, $n$ is the number of learnable prompt tokens, and $d^{\prime}$ is the dimension of the transformer layer. In all earlier works, including CLIP, the $\textbf{cls}_{\mathcal{I}}$ output token is passed through the final projection layer $proj\in \mathbb{R}^{d^{\prime}\times d}$ of $\theta$ to obtain the \textit{global image feature} $\theta(\mathbf{x})\in\mathbb{R}^{d}$. These features capture the global context of the image but may not capture local object-level semantics~\citep{denseclip}. 
We aim to utilize the rich part-level local information hidden in the $196$ patch tokens and establish their association with class descriptions. To obtain the \textit{local image features} $\theta^{l}(\mathbf{x})\in \mathbb{R}^{196\times d}$, we pass the patch tokens through $proj$ and add a learnable $d$-dimensional bias offset as shown in Fig.~\ref{bias_figure}. This bias is added to fine-tune $proj$ with local information, which otherwise is used only to obtain global image features from the last transformer block. 

We obtain description-guided image features by performing a parameter free cross-attention with class description features as queries, and local image features as both keys and values.
\begin{equation*}
    \theta^{desc}(\mathbf{x}) = CrossAttention(Q=\phi(A), K=\theta^{l}(\mathbf{x}), V=\theta^{l}(\mathbf{x}))
\end{equation*}
Here, $\phi(A)\in\mathbb{R}^{N\times d}$ are the class description features for all class descriptions $A$, $\theta^{l}(\mathbf{x})\in\mathbb{R}^{196\times d}$ are the local features of an image. The description-guided image features $\theta^{desc}(\mathbf{x})\in\mathbb{R}^{N\times d}$ encode part-level local information relevant to the $N$ descriptions. For any description, the cross-attention module computes a weighted combination of the $196$ local features, where the weights are determined by the similarity between the image patch and the description. Note that we obtain $N$ description features, one per description, for a single image. Since descriptions are common across classes and even datasets, these features contain information that can transfer to novel classes.

\vspace{3pt}
\noindent \textbf{Fusing Description-Guided Features with Global Image Feature.} The description-guided image features described above use class descriptions from all classes, and not just the ground-truth class of the image. Since the class descriptions generated by LLMs may be noisy, not all descriptions are relevant to a specific image. 
To address this, we introduce a \textit{relevance score} $\mathbf{r} \in [0,1]^N$, which quantifies each description's similarity to the image. This is computed as: \[\mathbf{r}=softmax(\phi(A) \cdot \theta(\mathbf{x}))\] 
We perform a weighted average of $\theta^{desc}(\mathbf{x})$ with $\mathbf{r}$, and obtain the \textit{mean description-guided} feature $\bar{\theta}^{desc}(\mathbf{x})\in\mathbb{R}^d$, which captures finer contexts in an image and is computed as: \[ \bar{\theta}^{desc}(\mathbf{x}) = \theta^{desc}(\mathbf{x})^{\intercal} \cdot \mathbf{r}\]



For an image, the global image feature $\theta(\mathbf{x})\in \mathbb{R}^{d}$ encodes class information pertaining to the image and the mean description-guided feature $\bar{\theta}^{desc}(\mathbf{x})\in \mathbb{R}^{d}$ encodes part-level visual context. We perform a fusion of both these features to yield the final \textit{fused image feature} $\hat{\theta}(\mathbf{x})$. 
\begin{equation*}
    \Hat{\theta}(\mathbf{x}) = (1-\alpha)\cdot \theta(\mathbf{x}) + \alpha\cdot \bar{\theta}^{desc}(\mathbf{x})
\end{equation*}
We give a higher weight $\alpha\in[0,1]$ to the part-level features $\bar{\theta}^{desc}(\mathbf{x})$ of an image if the descriptions attend strongly to specific patches of the image. This indicates the specificity of certain descriptions to some parts of the image. To see this consider the case of a background image. Clearly such an image is uninformative w.r.t any class description, and its part-level features can be discounted. For each description, the maximum attention weight over image patches is a proxy for the specificity of the description. We then define $\alpha$ as the average specificity for all descriptions. The fused image feature $\Hat{\theta}(\mathbf{x})\in\mathbb{R}^d$ contains global visual semantics as well as part-level semantics.  

\vspace{-7pt}
\subsection{Description-Guided Semantic Alignment}
\label{sec:aligning promted image and text features}
\vspace{-3pt}
Given an image $\mathbf{x}$, we obtain the fused image feature $\Hat{\theta}(\mathbf{x})$ as described in \S~\ref{sec:leveraging_descriptions_for_image_features}. For every class $y\in\mathcal{Y}$, we obtain the description-guided text features $\phi(y;A_y)$ as described in \S~\ref{sec:leveraging_descriptions_for_text_features}. We denote the learnable prompt vectors by $\rho$, and we represent prompted features with subscript $p$. For instance, the prompted fused image feature is $\Hat{\theta}_p(\mathbf{x})$, and so on. Prompts are trained by minimizing the negative log-likelihood of the training data $\{(\mathbf{x}_i, y_i)\}_{i=1}^B$: 
\vspace{-4pt}
\begin{align*}
    L_{ce}(\rho) &= -\frac{1}{B}\sum\limits_{i=1}^B \log \frac{\exp(\xi(\hat{\theta}_p(\mathbf{x_i}), \phi_p(y_i;A_{y_i}))/\tau)}{\sum\limits_{y\in\mathcal{Y}} \exp( \xi(\hat{\theta}_p(\mathbf{x_i}), \phi_p(y;A_{y}))/\tau)}\\
    \text{where }& \xi(\hat{\theta}_p(\mathbf{x}), \phi_p(y;A_y)) = \frac{1}{\abs{A_y}}\sum\limits_{a\in A_y} sim(\hat{\theta}_p(\mathbf{x}), \phi_p(y;a))
    \vspace{-2pt}
\end{align*}
where $\tau$ is the temperature parameter, and $sim$ is cosine similarity. To compute semantic alignment $\xi$, we aggregate similarity between the fused image feature and the description-guided text feature over all pertinent class descriptions and normalize by their count. A relevant description in the image enhances its similarity to the class; however, the absence of a description in the image does not penalize its similarity to the class. Following~\citep{kgcoop,promptsrc}, we add regularization terms designed to penalize prompted features that deviate significantly from their unprompted counterparts. We use the $L1$ penalty to regularize global image features and description guided text features.
\vspace{-4pt}
\begin{align*}
    L_{steer}^v(\rho) &= \frac{1}{B} \sum\limits_{i=1}^B \lVert \theta_p(\mathbf{x_i}) - \theta(\mathbf{x_i})  \rVert_{1}\\
    L_{steer}^t(\rho) &= \frac{1}{|\mathcal{Y}|}\sum\limits_{y\in\mathcal{Y}} \lVert \phi_p(y;A_y) - \phi(y;A_y) \rVert_{1}
\end{align*}
The final objective is ${\mathcal{L}(\rho)=L_{ce}(\rho) + \lambda_1 L_{steer}^v(\rho) + \lambda_2 L_{steer}^t(\rho)}$, where $\lambda_1$ \& $\lambda_2$ are hyperparameters.

\textbf{Inference:} Let $\mathcal{Y}^{\prime}$ be the inference time label space, and $A_z$ be the class descriptions of class $z\in\mathcal{Y}^{\prime}$. Using the learned prompt $p$, we compute the prompted fused image feature and the description-guided text features for all classes in $\mathcal{Y}^{\prime}$. The class with the highest semantic alignment $\xi(\hat{\theta}_p(\mathbf{x}^{\prime}), \phi_p(z;A_z))$ is then predicted as the final label. The overall algorithm of \methodname{} is presented in Appendix \S~\ref{app sec algorithm} .






\vspace{-10pt}
\section{Experiments and Results}
\label{sec experiments}
\vspace{-5pt}
In this section, we comprehensively evaluate the generalization performance of SAP on two newly proposed benchmarks --  (i) Generalized Zero-Shot Classification (GZS) and (ii) Out-of-Vocabulary Classification (OVC) and existing benchmarks (iii) Base-to-Novel Generalization (B2N) and (iv) Cross-Dataset Generalization.

\textbf{Proposed Evaluation Benchmarks:}

\noindent\textbf{(i) Generalized Zero-Shot Classification (GZS).} In GZS, the label space of a dataset is equally split into disjoint base and novel classes. Only a small number (e.g., 16-shot) of labeled samples from the base classes are available as training data. However, during evaluation, the classification label space is the union of base and novel classes. As explained in \S~\ref{sec:prelims}, zero-shot classification performance depends on the label space considered, and introducing the union of base and novel classes into the label space tests the bias of the fine-tuned model towards base classes.  Hence, we believe this benchmark is a more realistic measure of the generalization performance of VLM fine-tuning methods. Though this setting has existed in traditional zero-shot learning~\citep{traditionalzsl}, we introduce it back into the realm of VLM evaluation.

\noindent\textbf{(ii) Out-of-Vocabulary Classification (OVC).} VLMs require explicit class names to perform classification~\citep{clip}. This is a limitation for images whose label lies outside the VLM's vocabulary. OVC tests the ability of a VLM to classify truly novel images without explicitly using class names. During inference, all class names are replaced with the word \textit{`object'}, and the model is tested on it's ability to classify an image based on descriptions alone. For example, to classify a \textit{`Pikachu'} image, we just use the descriptions \textit{\{`has a yellow body', ..., `has round red cheeks'\}} and not the class name \textit{`Pikachu'}, hence the text template looks like \texttt{`a photo of an object, which has a yellow body'} etc. The model is fine-tuned on base classes, and evaluated on base and novel classes separately by removing all class names. 

\textbf{Existing Evaluation Benchmarks:}

\noindent\textbf{(iii) Base-to-Novel Generalization (B2N).} In this setting, following prior work \citep{coop,cocoop,maple,kgcoop,promptsrc,logoprompt}, the dataset is split into equal disjoint base and novel classes, and the model is fine-tuned on few-shot (16-shot) training split of the base classes. During evaluation, unlike GZS, the label space is constrained to either just the base classes or just the novel classes. The testing phase for B2N is thus separate for base and novel classes, whereas the GZS benchmark has a unified testing phase.

\noindent\textbf{(iv) Cross-Dataset Generalization.} In this setting, the model is fine-tuned on ImageNet~\citep{imagenet} and tested on the remaining datasets. This measures the ability of a VLM fine-tuning method to generalize to novel datasets.

\noindent\textbf{Baselines.} We compare \methodname{},  against state-of-the-art baselines, including very recent prompt-tuning methods (summarized in Tab.~\ref{tab:comparison}), such as CLIP~\citep{clip}, CoOp~\citep{coop}, VPT~\citep{vpt}, CoCoOp~\citep{cocoop}, ProDA~\citep{proda}, MaPLe~\citep{maple}, KgCoOp~\citep{kgcoop}, ProGrad~\citep{prograd}, PSRC~\citep{promptsrc} and LoGoPrompt~\citep{logoprompt}. We also compare against contemporary works that use external knowledge, such as KAPT~\citep{kapt}, CLIP-VDT~\citep{vdt_adapter} and CoPrompt~\citep{coprompt}. 

\noindent\textbf{Datasets.} We follow~\citep{coop,cocoop,maple,promptsrc} to evaluate our method on 11 image classification datasets of varying complexity. These datasets encompass diverse domains, including generic object datasets like ImageNet~\citep{imagenet} and Caltech101~\citep{caltech}; fine-grained datasets like Stanford Cars~\citep{stanfordcars}, OxfordPets~\citep{oxfordpets}, Flowers102~\citep{oxfordflowers}, Food101~\citep{food101}, FGVCAircraft~\citep{fgvcaircraft}; scene recognition dataset SUN397~\citep{sun397}; action recognition dataset UCF101~\citep{ucf101}; texture dataset DTD~\citep{dtd}, and satellite image dataset EuroSAT~\citep{eurosat}.


\noindent\textbf{Overview of Results.} We present average base class accuracy, novel class accuracy, and their harmonic mean across 11 datasets for the GZS, OVC, B2N,  and Cross-Dataset benchmarks in \S~\ref{sec:empirical_results} -- Tab.~\ref{gzs_table}, Fig.~\ref{classficiationwithoutclass}, Tab.~\ref{basetonew}, and Tab.~\ref{crossdataset} respectively. Dataset-wise expanded tables for all benchmarks, along with Domain Generalization and ResNet-50 backbone results are present in Appendix \S~\ref{app sec additional results}. In \S~\ref{sec:additional_results}, we show class activation maps to visualize image regions most relevant to a class description, where SAP demonstrates better localization capabilities.
We study the goodness of our design choices in \S~\ref{sec:ablations} and show that part-level semantic alignment between image and text features helps learning better prompts.
\subsection{Main Results}
\label{sec:empirical_results}
\noindent \textbf{(i) Generalized Zero-Shot Classification.} 
This newly proposed benchmark tests the ability of a method towards it bias to base classes and also it's generalization to novel classes within a dataset. We compare SAP against baselines and report the results in Tab.~\ref{gzs_table}. The metric gBase is the average accuracy of test images belonging to base classes when the label space is the set of all classes (union of base and novel classes). The metric gNovel is the average accuracy of test images belonging to novel classes when the label space is the set of all classes. gHM is the harmonic mean of the gBase and gNovel. SAP's ability to leverage descriptions helps in mitigating the bias towards base classes, resulting in good generalized novel class accuracy. We outperform a recent state-of-the-art method PSRC, achieving better results in 8 out of 11 datasets (see Appendix \S~\ref{app sec additional results}), with a $\mathbf{\textbf{+1.21\%}}$ margin in gHM averaged over all 11 datasets. Compared to the second-best method MaPLe, we have a significant margin of $\mathbf{\textbf{+3.25\%}}$ in average gHM, outperforming it on all 11 datasets. We don't report the results of ProDA, LoGoPrompt, and KAPT in this setting due to code unavailability.

\begin{table}
    \centering
    \footnotesize
    \scalebox{0.68}{
    \begin{tabular}{cc|ccccccccc|c}
    \toprule
         \textbf{Dataset}& &\textbf{CLIP}&\textbf{CoOp}&\textbf{VPT}&\textbf{CoCoOp}&\textbf{MaPLe}&\textbf{KgCoOp}&\textbf{ProGrad}&\textbf{PSRC}&\textbf{CLIP-VDT}&\textbf{SAP}\\
        &&(ICML '21)&(IJCV '22)&(ECCV '22)&(CVPR '22)&(CVPR '23)&(CVPR '23)&(ICCV '23)&(ICCV '23)&(ICCVW '23)&(Ours)\\
     \midrule
     \textbf{Average}& gBase&\large 60.81&\large 75.19& \large73.48&\large73.13&\large75.47&\large76.86&\large70.15&\large\underline{78.81}&\large63.75&\large\textbf{79.47 (\textcolor{darkgreen}{+0.66})}\\
     \textbf{on 11}&gNovel&\large63.21&\large60.39&\large66.62&\large65.23&\large67.09&\large62.12&\large55.07&\large\underline{68.13}&\large63.89&\large\textbf{69.75 (\textcolor{darkgreen}{+1.62})}\\
     \textbf{datasets}&gHM&\large61.99&\large66.99&\large69.89&\large68.96&\large71.04&\large68.71&\large61.70&\large\underline{73.08}&\large63.82&\large\textbf{74.29 (\textcolor{darkgreen}{+1.21})}\\
     \bottomrule
    \end{tabular}
    }
     \caption{Results on the GZS benchmark. gNovel \& gBase indicate the accuracy of the novel classes and base classes respectively under the joint classification label space. gHM is the harmonic mean of gBase and gNovel. The best numbers are in \textbf{bold}, and the second best are \underline{underlined}. SAP outperforms the best performing baseline on average gBase (by $+0.66\%$), gNovel (by $+1.62\%$), and gHM (by $+1.21$) computed across all datasets. Detailed dataset-wise results are presented in Appendix \S~\ref{app sec additional results}.}
     \vspace{-15pt}
     \label{gzs_table}
\end{table}

\begin{figure}
    \centering
    \includegraphics[width=0.9\textwidth]{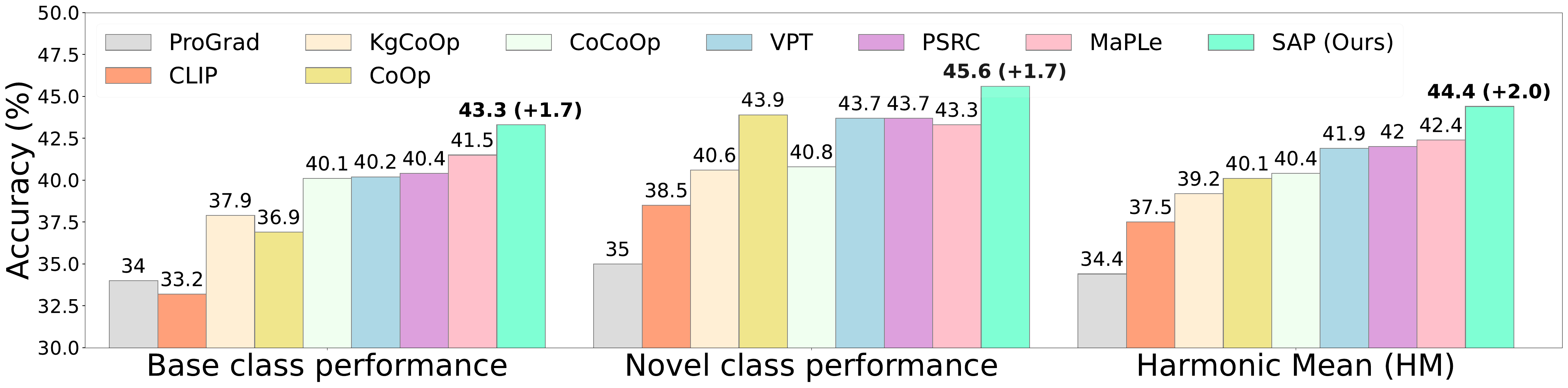}
    \caption{Comparison in the OVC setting. We show average Base, Novel, and HM accuracies over all 11 datasets. During evaluation, descriptions of each class are provided instead of the class name, and visual recognition is conducted based on these descriptions. SAP outperforms baselines by average Base (by $+1.75\%$), Novel (by $+1.76\%$) and HM (by $+2.04\%$) computed over all datasets. Detailed dataset-wise results are presented in Appendix \S~\ref{app sec additional results}.}
    \vspace{-5pt}
    \label{classficiationwithoutclass}
\end{figure}

\noindent\textbf{(ii) Out-of-Vocabulary Classification.} In this newly proposed benchmark, we study the ability of a pretrained VLM to classify images whose class names lie outside CLIP's vocabulary. Since the list of datasets CLIP was trained on is not public knowledge, to empirically evaluate this setting we use the standard 11 datasets itself, but remove access to class-names during evaluation. Similar to the B2N setting, all models are trained on base-class images. For all baselines (including ours), we find the similarity of an image $\textbf{x}$ with a class $y$ (not given to the model) as the average similarity between the image and the corresponding class-descriptions of $y$, which are known. We report average accuracies on 11 datasets in Fig.~\ref{classficiationwithoutclass}, where we outperform MaPLe~\citep{maple} by $\textbf{+2.04\%}$ in HM.

\noindent\textbf{(iii) Base-to-Novel Generalization.} In this setting, we compare our method with twelve baselines and report the average accuracies in Tab.~\ref{basetonew}, where we outperform all baselines. We report per dataset accuracies in the Appendix \S~\ref{app sec additional results}, and show that SAP outperforms the state-of-the-art method PSRC in 7 out of 11 datasets while retaining performance in the others. We show significant gains in challenging datasets such as EuroSAT and DTD, where we outperform PSRC by a margin of $\textbf{+5.66\%}$ and $\textbf{+2.92\%}$ in HM respectively. We also show a considerable boost in performance on the UCF-101 dataset, which contains a wide variety of human actions captured in diverse settings, where we show an improvement of $\textbf{+2.49\%}$ in HM over PSRC. These results indicate that SAP can integrate part-level knowledge provided by class descriptions to learn generalizable prompts.

\begin{table}
    \centering
    \footnotesize
    \scalebox{0.65}{
    \begin{tabular}{cc|cccccccccccc|c}
    \toprule
         \textbf{Dataset}& &\textbf{CLIP}&\textbf{CoOp}&\textbf{VPT}&\textbf{CoCoOp}&\textbf{ProDA}&\textbf{MaPLe}&\textbf{KgCoOp}&\textbf{ProGrad}&\textbf{PSRC}&\textbf{L.Prompt}&\textbf{CLIP-VDT}&\textbf{KAPT}&\textbf{SAP (Ours)}\\
     \midrule
         \multirow{3}{4em}{\textbf{Average on 11 datasets}} & Base &\large 69.34&\large 82.69&\large 80.81&\large 80.47&\large 81.56&\large 82.28&\large 80.73&\large 82.48&\large 84.26&\large \underline{84.47}&\large 82.48&\large 81.10&\large \textbf{84.68 (\textcolor{darkgreen}{+0.21})} \\
         & Novel &\large 74.22&\large 63.22&\large 70.36&\large 71.69&\large 72.30&\large 75.14&\large 73.60&\large 70.75&\large \underline{76.10}&\large 74.24&\large 74.50&\large 72.24&\large \textbf{77.51 (\textcolor{darkgreen}{+1.41})} \\
         & HM &\large 71.70&\large 71.66&\large 70.36&\large 75.83&\large 76.65&\large 78.55&\large 77.00&\large 76.16&\large \underline{79.97}&\large 79.03&\large 78.28&\large 76.41&\large \textbf{80.94 (\textcolor{darkgreen}{+0.97})} \\
    \bottomrule
    \end{tabular}
    }
    \caption{Comparison on Base-to-Novel Generalization benchmark. The best numbers are in \textbf{bold}, and the second best are \underline{underlined}. SAP outperforms the best performing baseline on average Base (by $+0.21\%$), Novel (by $+1.41\%$) and HM (by $+0.97\%$) computed over all datasets. Expanded tables are in Appendix \S~\ref{app sec additional results}.}
    \vspace{-10pt}
    \label{basetonew}
\end{table}

\vspace{4pt}

\noindent \textbf{(iv) Cross-Dataset Generalization.} We compare our method with nine baselines and outperform all of them as shown in Tab.~\ref{crossdataset}. SAP outperforms PSRC~\citep{promptsrc} by $\mathbf{\textbf{+1\%}}$ and MaPLe~\citep{maple} by $\mathbf{\textbf{+0.5\%}}$ on average test accuracy over all datasets, which indicates that our method learns prompts that generalize across datasets.

\begin{table}[H]
    \centering
    \vspace{7pt}
    \scalebox{0.73}{
    \begin{tabular}{c|ccccccccc|c}
    \toprule
         \textbf{Dataset}&\textbf{CoOp} &\textbf{CoCoOp}&\textbf{VPT}&\textbf{MaPLe}&\textbf{KgCoOp}&\textbf{ProGrad}&\textbf{PSRC}&\textbf{CLIP-VDT}&\textbf{KAPT}&\textbf{SAP (Ours)} \\
         \midrule
         Avg. on 10 Datasets&\large63.88&\large65.74&\large63.42&\large\underline{66.30}&\large65.49&\large57.36&\large65.81&\large53.98&\large61.50&\large\textbf{66.85 (\large\textcolor{darkgreen}{+0.55})}\\
         \bottomrule
    \end{tabular}
    }
    \caption{Cross-Dataset Generalization. Models are trained on Imagenet and tested on the entire label space of new datasets without fine-tuning. SAP outperforms all baselines on average (see Appendix \S~\ref{app sec additional results}).}
    \label{crossdataset}
\end{table}

\noindent\textbf{Comparison against a recent method that uses external knowledge.} In Tab.~\ref{tab:attrmethods} we compare SAP against CoPrompt~\citep{coprompt} on the B2N benchmark. 
\begin{wraptable}[9]{r}{0.55\textwidth}
    \centering
    \footnotesize
    \scalebox{0.73}{
    \begin{tabular}{cc|cc|c}
    \toprule
&&\textbf{CoPrompt}&\textbf{CoPrompt*}&\textbf{SAP (Ours)}\\
         &&prompts+adapter& prompts&prompts\\
     \midrule
         \multirow{3}{4em}{\textbf{Average on 11 datasets}} & \large Base &\large84.00&\large83.40&\large\textbf{84.68 \large(\textcolor{darkgreen}{+1.28})} \\
         & \large Novel&\large77.23&\large76.90&\large\textbf{77.51 \large(\textcolor{darkgreen}{+0.61})} \\
         & \large HM &\large80.48&\large80.02&\large\textbf{80.94 \large(\textcolor{darkgreen}{+0.92})} \\
    \bottomrule
    \end{tabular}
}
    \caption{B2N results comparison against a recent method CoPrompt. SAP outnumbers the prompt-only version by a margin on Base (by +1.28\%), Novel (by +0.61\%), and HM (by +0.92\%).}
    \label{tab:attrmethods}
\end{wraptable}
CoPrompt is a recent work that uses class descriptions to tune prompts and adapters, with a total of $4.74M$ additional parameters over CLIP. SAP outperforms CoPrompt by $\textbf{+0.46\%}$ average HM, despite only having $36K$ additional learnable parameters over CLIP. We also compare SAP against a prompt-only version of CoPrompt, as indicated by CoPrompt* in Tab.~\ref{tab:attrmethods}, in which we outperform by $\textbf{+0.92\%}$ in average HM.
\subsection{Qualitative Results}
\label{sec:additional_results}


\begin{figure}
    \centering
    \includegraphics[width=0.95\textwidth]{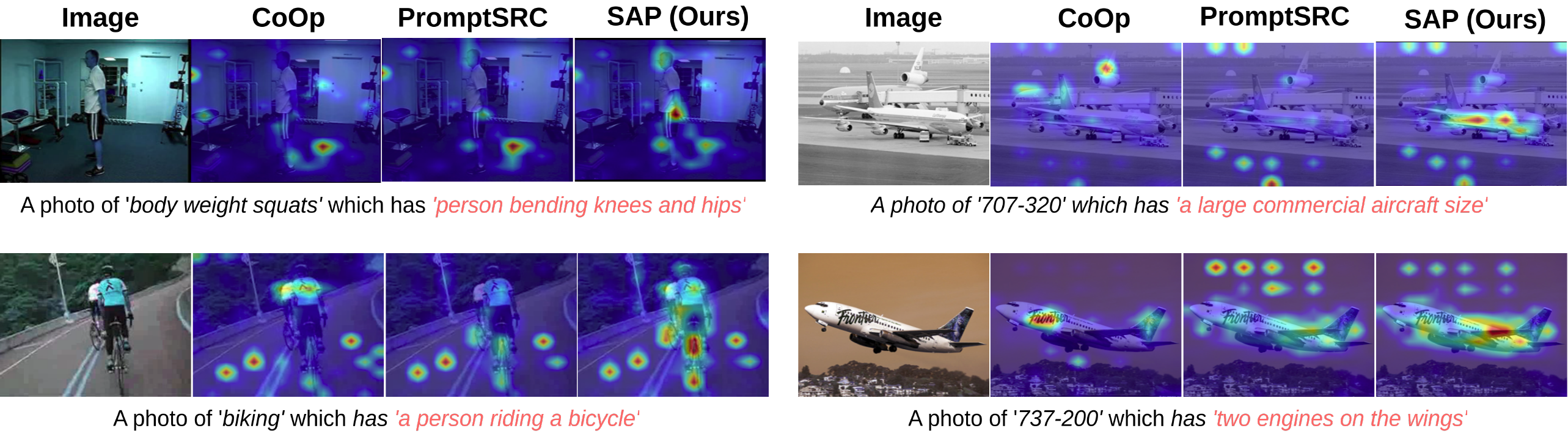}
    \caption{Images are highlighted at regions of highest activation relevant to specific text phrases, as identified by their prompted image and text encoders. Qualitatively, \methodname{} localizes better than the existing baselines.}
    \vspace{-15pt}
    \label{fig:saliency}
\end{figure}

\noindent\textbf{Class Activation Maps.}
We present Class Activation Maps (CAMs) for the ViT-backbone CLIP image encoder to show image regions that most correlate to a given text description. We visualize activations of the pre-final self-attention layer of the transformer that maximize the cosine similarity between an image and a given text description. We present qualitative results in Fig.~\ref{fig:saliency}, where prompts learned by our method lead to better localizations. We also propose an occlusion metric to measure the localization capabilities of our learned prompts. Given a description, we mask out parts of the image which are most activated w.r.t. the description. The occluded image is then classified by the pre-trained CLIP model. A class activation map localizes the description well if occluding image regions with the highest activations leads to the greatest drop in accuracy.

\begin{table}[H]
    \centering
    \scalebox{0.72}{
\begin{tabular}{lccccccc}\\
\toprule
\textbf{Method}&\textbf{Archery}&\textbf{Baby Crawling}&\textbf{Band Marching}&\textbf{Apply Eye Makeup}&\textbf{Apply Lipstick}&\textbf{Biking}&\textbf{Body Weight Squats}\\
\midrule
CoOp&\large57.39&\large64.42&\large61.99&\large75.00&\large78.66&\large55.15&\large53.97\\
PSRC&\large47.87&\large53.69&\large54.29&\large50.00&\large69.33&\large50.35&\large50.72\\
Ours&\large\textbf{44.34}&\large\textbf{49.66}&\large\textbf{51.58}&\large\textbf{40.90}&\large\textbf{62.66}&\large\textbf{47.96}&\large\textbf{48.73}\\
\midrule
&\textbf{707-320}&\textbf{747-200}&\textbf{737-200}&\textbf{727-200}&\textbf{C-130}&\textbf{CRJ-200}&\textbf{Boeing-717}\\
\midrule
CoOp&\large15.21&\large11.82&\large23.47&\large6.13&\large75.81&\large38.22&\large20.63\\
PSRC&\large6.14&\large8.84&\large21.42&\large3.06&\large75.86&\large32.45&\large23.58\\
Ours&\large\textbf{3.00}&\large\textbf{5.92}&\large\textbf{15.30}&\large\textbf{0.00}&\large\textbf{60.61}&\large\textbf{26.58}&\large\textbf{14.72}\\
\bottomrule
\end{tabular}}
\caption{Occlusion benchmark (lower number is better): Images are masked at regions of highest activation relevant to a given class description, as identified by prompted image and text encoders, and then evaluated using the pre-trained CLIP model. The lower the accuracy, the better are the localizations. We show results for a few specific classes from the UCF101 dataset (top) and FGVC-Aircraft dataset (bottom). For example, for the class \textit{`body weight squats'}, we use the description \textit{`person bending knees and hips'}.}
\label{tab:my_label}
\end{table}

For instance, for the text phrase \textit{`a photo of a 737-200, which has two engines on the wings'} we find that masking out important regions given by our prompted image encoder leads to an accuracy of $15.30\%$. This drop is higher than that of PSRC, whose accuracy drops only to $21.42\%$. This suggests that regions which are deemed important by \methodname{} are highly correlated to the text phrase. Our parameter-free cross-attention module helps us learn prompts that focus on part-level image information.

\vspace{-5pt}
\subsection{Ablation Studies}
\label{sec:ablations}
\vspace{-5pt}

\noindent \textbf{Study on Design Choices.} In this section we justify our design choice of \textit{computing semantic alignment as the average similarity between the fused image feature and various description-guided text features}. Our key contribution is not just integrating descriptions into prompt learning for VLMs, but \textit{how} descriptions are integrated into \textit{both} visual and text modalities. We consider three alternative ways to incorporate class descriptions and show that our methodology leads to the best results. For our first alternative, we show that taking the \textit{unnormalized mean} of description-guided text features to compute similarity leads to a drop in performance (SAP w/ mean text feature in Tab.~\ref{tab:text_analysis}). That is, computing semantic alignment as $\xi(\hat{\theta}_p(\mathbf{x}), \phi_p(y;A_y)) = sim(\hat{\theta}_p(\mathbf{x}),\frac{1}{\abs{A_y}}\sum\limits_{a\in A_y} \phi_p(y;a))$, leads to a drop in performance. This is in contrast to our design choice of taking the \textit{mean} similarity, as shown in Eq.~\ref{eq:semantic alignment}. Intuitively, descriptions of a class that are not well represented in pre-trained CLIP result in description-guided features with a low norm because CLIP has not encountered such associations during training. Information related to such descriptions is lost when the description-guided features are simply averaged out, without normalization.

\begin{wraptable}[10]{r}{0.45\textwidth}
\centering
\begin{tabular}{lc}
\toprule
\textbf{Method}&\textbf{Avg HM}\\
\midrule
SAP&\textbf{\large 80.94}\\
SAP w/ mean text feature&\large 80.31\\
SAP w/ agg descriptions&\large 79.17\\
CLIP-VDT Text + SAP's Visual&\large 78.63\\
\bottomrule
\end{tabular}
\caption{Comparison with alternative design choices for incorporating class descriptions into the text modality.}
\label{tab:text_analysis}
\end{wraptable}
We also observe that simply appending all class descriptions at once to generate a single description-guided text feature also leads to a drop in performance (SAP w/ agg descriptions in Tab.~\ref{tab:text_analysis}). Finally we show that replacing our text modality construction with that used by CLIP-VDT (CLIP-VDT text + SAP's Visual in Tab.~\ref{tab:text_analysis}) leads to a significant drop in average HM. These experiments show that how we add class descriptions is important, and that our approach is different from recent approaches that uses external information. We show average HM results across all 11 datasets of other design choices in Tab.~\ref{tab:text_analysis}.

\noindent\textbf{Effect of Removing Learnable Bias.} To study the effect of adding a learnable bias to obtain local features, we conduct an ablation study. Tab.~\ref{tab:ablations} shows that adding a bias is a parameter-efficient way to learn good local image features.

\begin{wraptable}[15]{r}{0.56\textwidth}
    \centering
    \scalebox{0.8}{
    \begin{tabular}{l|ccc}
    \toprule
        \textbf{Method} & \textbf{Avg. Base}&\textbf{Avg. Novel}&\textbf{Avg. HM}\\
         \midrule
         \multicolumn{4}{c}{Effect of Removing Learnable Bias}\\
         \midrule
         SAP w/o bias & 84.55 & 75.72 & 79.9\\
         SAP & \textbf{84.68} & \textbf{77.51} & \textbf{80.94}\\
         \midrule
         \multicolumn{4}{c}{Effect of Removing Class descriptions from the Text Modality}\\
         \midrule
         SAP - TG & 84.62 & 74.79 & 79.41\\
         SAP & \textbf{84.68} & \textbf{77.51} & \textbf{80.94}\\
         \midrule
         \multicolumn{4}{c}{Effect of Removing Class Descriptions from the Image Modality}\\
         \midrule
         SAP w/ global & 84.56& 77.04 & 80.63\\
         SAP w/ global \& local & 84.66& 76.81 & 80.55\\
         SAP & \textbf{84.68} & \textbf{77.51} & \textbf{80.94}\\
         \bottomrule
    \end{tabular}
    }
    \caption{All results are on the B2N generalization benchmark, and are average results over 11 datasets.}
    \label{tab:ablations}
\end{wraptable}
\noindent\textbf{Effect of Removing Class Descriptions.} Our method SAP incorporates class descriptions in both image and text modalities, as described in \S~\ref{sec:leveraging_descriptions_for_text_features} \& \S~\ref{sec:leveraging_descriptions_for_image_features}. Here we study the effect of removing description guidance from both modalities. To remove description guidance from text, we just use the default class name template i.e. \texttt{`a photo of a [y]'}, without using any class description. We denote this baseline as SAP-TG. The results shown in Tab.~\ref{tab:ablations} indicate that adding class descriptions to the text modality, as SAP does, helps a lot. To study the effect of removing class descriptions from the image modality, we construct baselines by removing the cross attention module. We first consider a baseline that uses just the global image feature $\theta(x)$ instead of the fused feature and call this SAP w/ global. Then, we consider a baseline that naively combines global and local features (without incorporating class descriptions via cross-attention) by averaging them and denote it by SAP w/ global \& local. Note that both baselines construct description guided text features. The results presented in Tab.~\ref{tab:ablations} justify our design choice of incorporating class descriptions into the image modality. Furthermore our method to incorporate class descriptions into images is through a fully non-parametric cross-attention, and adds little computational overhead.
\vspace{-7pt}
\section{Conclusions}
\vspace{-7pt}
Prompt learning has emerged as a valuable technique for fine-tuning VLMs for downstream tasks. However, existing methods encounter challenges such as overfitting due to limited training data and difficulties handling larger label spaces during evaluation, resulting in bias towards seen classes. Additionally, these methods struggle when class labels are not present in the vocabulary. We study if better text semantics can improve prompt learning, and propose an approach, named SAP, that learns prompts which better generalize to novel classes. Our proposed approach highlights that careful part-level semantic alignment between image and text features is crucial to leverage additional semantic information. We showcase the efficacy of our approach across four benchmarks, demonstrating significant improvements. We hope this work inspires further exploration into leveraging class descriptions in VLMs.

\bibliography{main}

\begin{thebibliography}{43}
\providecommand{\natexlab}[1]{#1}
\providecommand{\url}[1]{\texttt{#1}}
\expandafter\ifx\csname urlstyle\endcsname\relax
  \providecommand{\doi}[1]{doi: #1}\else
  \providecommand{\doi}{doi: \begingroup \urlstyle{rm}\Url}\fi

\bibitem[Bossard et~al.(2014)Bossard, Guillaumin, and Gool]{food101}
Lukas Bossard, Matthieu Guillaumin, and Luc~Van Gool.
\newblock Food-101 - mining discriminative components with random forests.
\newblock In \emph{European Conference on Computer Vision}, 2014.

\bibitem[Cimpoi et~al.(2013)Cimpoi, Maji, Kokkinos, Mohamed, and Vedaldi]{dtd}
Mircea Cimpoi, Subhransu Maji, Iasonas Kokkinos, Sammy Mohamed, and Andrea Vedaldi.
\newblock Describing textures in the wild.
\newblock \emph{2014 IEEE Conference on Computer Vision and Pattern Recognition}, pp.\  3606--3613, 2013.

\bibitem[Deng et~al.(2009)Deng, Dong, Socher, Li, Li, and Fei-Fei]{imagenet}
Jia Deng, Wei Dong, Richard Socher, Li-Jia Li, K.~Li, and Li~Fei-Fei.
\newblock Imagenet: A large-scale hierarchical image database.
\newblock \emph{2009 IEEE Conference on Computer Vision and Pattern Recognition}, pp.\  248--255, 2009.

\bibitem[Dosovitskiy et~al.(2021)Dosovitskiy, Beyer, Kolesnikov, Weissenborn, Zhai, Unterthiner, Dehghani, Minderer, Heigold, Gelly, Uszkoreit, and Houlsby]{dosovitskiy2021an}
Alexey Dosovitskiy, Lucas Beyer, Alexander Kolesnikov, Dirk Weissenborn, Xiaohua Zhai, Thomas Unterthiner, Mostafa Dehghani, Matthias Minderer, Georg Heigold, Sylvain Gelly, Jakob Uszkoreit, and Neil Houlsby.
\newblock An image is worth 16x16 words: Transformers for image recognition at scale.
\newblock In \emph{International Conference on Learning Representations}, 2021.

\bibitem[Fahes et~al.(2023)Fahes, Vu, Bursuc, P{\'e}rez, and de~Charette]{poda}
Mohammad Fahes, Tuan-Hung Vu, Andrei Bursuc, Patrick P{\'e}rez, and Raoul de~Charette.
\newblock P{\o}da: Prompt-driven zero-shot domain adaptation.
\newblock In \emph{ICCV}, 2023.

\bibitem[Fei-Fei et~al.(2004)Fei-Fei, Fergus, and Perona]{caltech}
Li~Fei-Fei, Rob Fergus, and Pietro Perona.
\newblock Learning generative visual models from few training examples: An incremental bayesian approach tested on 101 object categories.
\newblock \emph{2004 Conference on Computer Vision and Pattern Recognition Workshop}, pp.\  178--178, 2004.

\bibitem[Feng et~al.(2022)Feng, Zhong, Jie, Chu, Ren, Wei, Xie, and Ma]{feng2022promptdet}
Chengjian Feng, Yujie Zhong, Zequn Jie, Xiangxiang Chu, Haibing Ren, Xiaolin Wei, Weidi Xie, and Lin Ma.
\newblock Promptdet: Towards open-vocabulary detection using uncurated images.
\newblock In \emph{Proceedings of the European Conference on Computer Vision}, 2022.

\bibitem[Hagendorff et~al.(2022)Hagendorff, Fabi, and Kosinski]{gpt35}
Thilo Hagendorff, Sarah Fabi, and Michal Kosinski.
\newblock Machine intuition: Uncovering human-like intuitive decision-making in gpt-3.5, 12 2022.

\bibitem[He et~al.(2015)He, Zhang, Ren, and Sun]{resnet}
Kaiming He, X.~Zhang, Shaoqing Ren, and Jian Sun.
\newblock Deep residual learning for image recognition.
\newblock \emph{2016 IEEE Conference on Computer Vision and Pattern Recognition (CVPR)}, pp.\  770--778, 2015.

\bibitem[Helber et~al.(2017)Helber, Bischke, Dengel, and Borth]{eurosat}
Patrick Helber, Benjamin Bischke, Andreas~R. Dengel, and Damian Borth.
\newblock Eurosat: A novel dataset and deep learning benchmark for land use and land cover classification.
\newblock \emph{IEEE Journal of Selected Topics in Applied Earth Observations and Remote Sensing}, 12:\penalty0 2217--2226, 2017.

\bibitem[Jia et~al.(2021)Jia, Yang, Xia, Chen, Parekh, Pham, Le, Sung, Li, and Duerig]{align}
Chao Jia, Yinfei Yang, Ye~Xia, Yi-Ting Chen, Zarana Parekh, Hieu Pham, Quoc Le, Yun-Hsuan Sung, Zhen Li, and Tom Duerig.
\newblock Scaling up visual and vision-language representation learning with noisy text supervision.
\newblock In \emph{International conference on machine learning}, pp.\  4904--4916. PMLR, 2021.

\bibitem[Jia et~al.(2022)Jia, Tang, Chen, Cardie, Belongie, Hariharan, and Lim]{vpt}
Menglin Jia, Luming Tang, Bor-Chun Chen, Claire Cardie, Serge~J. Belongie, Bharath Hariharan, and Ser~Nam Lim.
\newblock Visual prompt tuning.
\newblock \emph{ArXiv}, abs/2203.12119, 2022.

\bibitem[Kan et~al.(2023)Kan, Wang, Lu, Zhen, Guan, and Zheng]{kapt}
Baoshuo Kan, Teng Wang, Wenpeng Lu, Xiantong Zhen, Weili Guan, and Feng Zheng.
\newblock Knowledge-aware prompt tuning for generalizable vision-language models.
\newblock \emph{2023 IEEE/CVF International Conference on Computer Vision (ICCV)}, pp.\  15624--15634, 2023.

\bibitem[Khattak et~al.(2022)Khattak, Rasheed, Maaz, Khan, and Khan]{maple}
Muhammad~Uzair Khattak, Hanoona Rasheed, Muhammad Maaz, Salman Khan, and Fahad~Shahbaz Khan.
\newblock Maple: Multi-modal prompt learning.
\newblock \emph{2023 IEEE/CVF Conference on Computer Vision and Pattern Recognition (CVPR)}, pp.\  19113--19122, 2022.

\bibitem[Khattak et~al.(2023)Khattak, Wasim, Naseer, Khan, Yang, and Khan]{promptsrc}
Muhammad~Uzair Khattak, Syed~Talal Wasim, Muzammal Naseer, Salman Khan, Ming-Hsuan Yang, and Fahad~Shahbaz Khan.
\newblock Self-regulating prompts: Foundational model adaptation without forgetting.
\newblock In \emph{Proceedings of the IEEE/CVF International Conference on Computer Vision (ICCV)}, pp.\  15190--15200, October 2023.

\bibitem[Krause et~al.(2013)Krause, Stark, Deng, and Fei-Fei]{stanfordcars}
Jonathan Krause, Michael Stark, Jia Deng, and Li~Fei-Fei.
\newblock 3d object representations for fine-grained categorization.
\newblock \emph{2013 IEEE International Conference on Computer Vision Workshops}, pp.\  554--561, 2013.

\bibitem[Lampert et~al.(2009)Lampert, Nickisch, and Harmeling]{awa}
Christoph~H Lampert, Hannes Nickisch, and Stefan Harmeling.
\newblock Learning to detect unseen object classes by between-class attribute transfer.
\newblock In \emph{IEEE Conference on Computer Vision and Pattern Recognition (CVPR)}, pp.\  951--958. IEEE, 2009.

\bibitem[Liang et~al.(2022)Liang, Wu, Dai, Li, Zhao, Zhang, Zhang, Vajda, and Marculescu]{openvocab}
Feng Liang, Bichen Wu, Xiaoliang Dai, Kunpeng Li, Yinan Zhao, Hang Zhang, Peizhao Zhang, P{\'e}ter Vajda, and Diana Marculescu.
\newblock Open-vocabulary semantic segmentation with mask-adapted clip.
\newblock \emph{2023 IEEE/CVF Conference on Computer Vision and Pattern Recognition (CVPR)}, pp.\  7061--7070, 2022.

\bibitem[Liu et~al.(2023)Liu, Li, Zhang, Wei, Bai, and Zhao]{zslcvpr23}
Man Liu, Feng Li, Chunjie Zhang, Yunchao Wei, Huihui Bai, and Yao Zhao.
\newblock Progressive semantic-visual mutual adaption for generalized zero-shot learning.
\newblock \emph{2023 IEEE/CVF Conference on Computer Vision and Pattern Recognition (CVPR)}, pp.\  15337--15346, 2023.

\bibitem[Lu et~al.(2022)Lu, Liu, Zhang, Liu, and Tian]{proda}
Yuning Lu, Jianzhuang Liu, Yonggang Zhang, Yajing Liu, and Xinmei Tian.
\newblock Prompt distribution learning.
\newblock \emph{2022 IEEE/CVF Conference on Computer Vision and Pattern Recognition (CVPR)}, pp.\  5196--5205, 2022.

\bibitem[L\"uddecke \& Ecker(2022)L\"uddecke and Ecker]{segmentation}
Timo L\"uddecke and Alexander Ecker.
\newblock Image segmentation using text and image prompts.
\newblock In \emph{Proceedings of the IEEE/CVF Conference on Computer Vision and Pattern Recognition (CVPR)}, pp.\  7086--7096, June 2022.

\bibitem[Maji et~al.(2013)Maji, Rahtu, Kannala, Blaschko, and Vedaldi]{fgvcaircraft}
Subhransu Maji, Esa Rahtu, Juho Kannala, Matthew~B. Blaschko, and Andrea Vedaldi.
\newblock Fine-grained visual classification of aircraft.
\newblock \emph{ArXiv}, abs/1306.5151, 2013.

\bibitem[Maniparambil et~al.(2023)Maniparambil, Vorster, Molloy, Murphy, McGuinness, and O'Connor]{vdt_adapter}
Mayug Maniparambil, Chris Vorster, Derek Molloy, Noel Murphy, Kevin McGuinness, and Noel~E. O'Connor.
\newblock Enhancing clip with gpt-4: Harnessing visual descriptions as prompts.
\newblock \emph{2023 IEEE/CVF International Conference on Computer Vision Workshops (ICCVW)}, pp.\  262--271, 2023.

\bibitem[Menon \& Vondrick(2023)Menon and Vondrick]{iclr23}
Sachit Menon and Carl Vondrick.
\newblock Visual classification via description from large language models.
\newblock \emph{ICLR}, 2023.

\bibitem[Nilsback \& Zisserman(2008)Nilsback and Zisserman]{oxfordflowers}
Maria-Elena Nilsback and Andrew Zisserman.
\newblock Automated flower classification over a large number of classes.
\newblock \emph{2008 Sixth Indian Conference on Computer Vision, Graphics \& Image Processing}, pp.\  722--729, 2008.

\bibitem[Parkhi et~al.(2012)Parkhi, Vedaldi, Zisserman, and Jawahar]{oxfordpets}
Omkar~M. Parkhi, Andrea Vedaldi, Andrew Zisserman, and C.~V. Jawahar.
\newblock Cats and dogs.
\newblock \emph{2012 IEEE Conference on Computer Vision and Pattern Recognition}, pp.\  3498--3505, 2012.

\bibitem[Pratt et~al.(2022)Pratt, Liu, and Farhadi]{Pratt2022WhatDA}
Sarah Pratt, Rosanne Liu, and Ali Farhadi.
\newblock What does a platypus look like? generating customized prompts for zero-shot image classification.
\newblock \emph{2023 IEEE/CVF International Conference on Computer Vision (ICCV)}, pp.\  15645--15655, 2022.
\newblock URL \url{https://api.semanticscholar.org/CorpusID:252111028}.

\bibitem[Radford et~al.(2021)Radford, Kim, Hallacy, Ramesh, Goh, Agarwal, Sastry, Askell, Mishkin, Clark, Krueger, and Sutskever]{clip}
Alec Radford, Jong~Wook Kim, Chris Hallacy, Aditya Ramesh, Gabriel Goh, Sandhini Agarwal, Girish Sastry, Amanda Askell, Pamela Mishkin, Jack Clark, Gretchen Krueger, and Ilya Sutskever.
\newblock Learning transferable visual models from natural language supervision.
\newblock In \emph{International Conference on Machine Learning}, 2021.

\bibitem[Rao et~al.(2021)Rao, Zhao, Chen, Tang, Zhu, Huang, Zhou, and Lu]{denseclip}
Yongming Rao, Wenliang Zhao, Guangyi Chen, Yansong Tang, Zheng Zhu, Guan Huang, Jie Zhou, and Jiwen Lu.
\newblock Denseclip: Language-guided dense prediction with context-aware prompting.
\newblock \emph{2022 IEEE/CVF Conference on Computer Vision and Pattern Recognition (CVPR)}, pp.\  18061--18070, 2021.

\bibitem[Rasheed et~al.(2022)Rasheed, Khattak, Maaz, Khan, and Khan]{ivlp}
Hanoona Rasheed, Muhammad~Uzair Khattak, Muhammad Maaz, Salman Khan, and Fahad~Shahbaz Khan.
\newblock Fine-tuned clip models are efficient video learners.
\newblock \emph{2023 IEEE/CVF Conference on Computer Vision and Pattern Recognition (CVPR)}, pp.\  6545--6554, 2022.

\bibitem[Roy \& Etemad(2024)Roy and Etemad]{coprompt}
Shuvendu Roy and Ali Etemad.
\newblock Consistency-guided prompt learning for vision-language models.
\newblock In \emph{The Twelfth International Conference on Learning Representations}, 2024.

\bibitem[Selvaraju et~al.(2017)Selvaraju, Cogswell, Das, Vedantam, Parikh, and Batra]{gradcam}
Ramprasaath~R. Selvaraju, Michael Cogswell, Abhishek Das, Ramakrishna Vedantam, Devi Parikh, and Dhruv Batra.
\newblock Grad-cam: Visual explanations from deep networks via gradient-based localization.
\newblock In \emph{2017 IEEE International Conference on Computer Vision (ICCV)}, pp.\  618--626, 2017.

\bibitem[Shi \& Yang(2023)Shi and Yang]{logoprompt}
Cheng Shi and Sibei Yang.
\newblock Logoprompt: Synthetic text images can be good visual prompts for vision-language models.
\newblock In \emph{Proceedings of the IEEE/CVF International Conference on Computer Vision (ICCV)}, pp.\  2932--2941, October 2023.

\bibitem[Soomro et~al.(2012)Soomro, Zamir, and Shah]{ucf101}
Khurram Soomro, Amir~Roshan Zamir, and Mubarak Shah.
\newblock Ucf101: A dataset of 101 human actions classes from videos in the wild.
\newblock \emph{ArXiv}, abs/1212.0402, 2012.

\bibitem[van~den Oord et~al.(2018)van~den Oord, Li, and Vinyals]{Oord2018RepresentationLW}
A{\"a}ron van~den Oord, Yazhe Li, and Oriol Vinyals.
\newblock Representation learning with contrastive predictive coding.
\newblock \emph{ArXiv}, abs/1807.03748, 2018.

\bibitem[Wah et~al.(2011)Wah, Branson, Welinder, Perona, and Belongie]{cub}
C.~Wah, Steve Branson, P.~Welinder, P.~Perona, and Serge~J. Belongie.
\newblock The caltech-ucsd birds-200-2011 dataset.
\newblock In \emph{.} California Institute of Technology, 2011.

\bibitem[Xian et~al.(2017)Xian, Schiele, and Akata]{traditionalzsl}
Yongqin Xian, Bernt Schiele, and Zeynep Akata.
\newblock Zero-shot learning — the good, the bad and the ugly.
\newblock \emph{2017 IEEE Conference on Computer Vision and Pattern Recognition (CVPR)}, pp.\  3077--3086, 2017.

\bibitem[Xiao et~al.(2010)Xiao, Hays, Ehinger, Oliva, and Torralba]{sun397}
Jianxiong Xiao, James Hays, Krista~A. Ehinger, Aude Oliva, and Antonio Torralba.
\newblock Sun database: Large-scale scene recognition from abbey to zoo.
\newblock \emph{2010 IEEE Computer Society Conference on Computer Vision and Pattern Recognition}, pp.\  3485--3492, 2010.

\bibitem[Yang et~al.(2022)Yang, Panagopoulou, Zhou, Jin, Callison-Burch, and Yatskar]{labo}
Yue Yang, Artemis Panagopoulou, Shenghao Zhou, Daniel Jin, Chris Callison-Burch, and Mark Yatskar.
\newblock Language in a bottle: Language model guided concept bottlenecks for interpretable image classification.
\newblock \emph{2023 IEEE/CVF Conference on Computer Vision and Pattern Recognition (CVPR)}, pp.\  19187--19197, 2022.

\bibitem[Yao et~al.(2023)Yao, Zhang, and Xu]{kgcoop}
Hantao Yao, Rui Zhang, and Changsheng Xu.
\newblock Visual-language prompt tuning with knowledge-guided context optimization.
\newblock \emph{2023 IEEE/CVF Conference on Computer Vision and Pattern Recognition (CVPR)}, pp.\  6757--6767, 2023.

\bibitem[Zhou et~al.(2021)Zhou, Yang, Loy, and Liu]{coop}
Kaiyang Zhou, Jingkang Yang, Chen~Change Loy, and Ziwei Liu.
\newblock Learning to prompt for vision-language models.
\newblock \emph{International Journal of Computer Vision}, 130:\penalty0 2337 -- 2348, 2021.

\bibitem[Zhou et~al.(2022)Zhou, Yang, Loy, and Liu]{cocoop}
Kaiyang Zhou, Jingkang Yang, Chen~Change Loy, and Ziwei Liu.
\newblock Conditional prompt learning for vision-language models.
\newblock \emph{2022 IEEE/CVF Conference on Computer Vision and Pattern Recognition (CVPR)}, pp.\  16795--16804, 2022.

\bibitem[Zhu et~al.(2023)Zhu, Niu, Han, Wu, and Zhang]{prograd}
Beier Zhu, Yulei Niu, Yucheng Han, Yue Wu, and Hanwang Zhang.
\newblock Prompt-aligned gradient for prompt tuning.
\newblock In \emph{Proceedings of the IEEE/CVF International Conference on Computer Vision (ICCV)}, pp.\  15659--15669, October 2023.

\end{thebibliography}
\bibliographystyle{tmlr}

\clearpage
\appendix
\section*{Appendix}
In this appendix, we present the following details.
\begin{itemize}[leftmargin=*]
\setlength \itemsep{-0.1em}
    \item List of notations used in this paper and their descriptions are in \S~\ref{app summary of noations}.
    \item Overall algorithm of \methodname{} is presented in \S~\ref{app sec algorithm}.
    \item Implementation details are in \S~\ref{app sec experimental setup}.
    \item Expanded dataset-wise tables, and additional experiments are presented in \S~\ref{app sec additional results}.
    \item Examples of class descriptions generated using GPT-3.5 are presented in \S~\ref{app sec attribute priors}.
    \item Limitations and Broader Impact in \S~\ref{sec:limitations}.
\end{itemize}
\section{Summary of Notations and Terminology}

\label{app summary of noations}
We use $\cdot$ (\textit{dot}) to represent various types of multiplication operations -- matrix multiplication, matrix-vector or vector-matrix product, and vector dot-product. Detailed descriptions of notations are presented in Tab.~\ref{notationstable}.

\begin{table}[H]
\vspace{5pt}
    \centering
    \begin{tabular}{c|l|l}
    \toprule
\textbf{Notation}&\textbf{Description}&\textbf{Dimension}\\
    \midrule
        $\mathcal{\theta}$& \multicolumn{2}{l}{Image Encoder}\\
        $\mathcal{\phi}$& \multicolumn{2}{l}{Text Encoder}\\
        $\mathcal{Y}$& \multicolumn{2}{l}{Classification label space}\\
        $\rho$& \multicolumn{2}{l}{Set of all learnable text and visual prompts}\\
        $B$&\multicolumn{2}{l}{Batch size}\\
        $N$&\multicolumn{2}{l}{Size of the set of descriptions}\\
        $n$&\multicolumn{2}{l}{Number of the learnable prompt tokens}\\
        $d$&\multicolumn{2}{l}{Dimension of the multimodal space}\\
        $A_y$& \multicolumn{2}{l}{LLM generated descriptions for class $y$}\\
        $A$& \multicolumn{2}{l}{Union of all descriptions
        of the classification label space}\\
        \midrule
        $\phi(A)$& Class descriptions features& $\mathbb{R}^{N\times d}$\\
        $\phi(y;A_{y})$& Description-guided text features of class $y$&$\mathbb{R}^{N\times d}$\\
        $\theta(x)$&Global image feature & $\mathbb{R}^{d}$\\
        $\theta^{l}(x)$&Local image feature & $\mathbb{R}^{M\times d}$\\
        $\theta^{desc}(x)$&Description-guided image features&$\mathbb{R}^{N\times d}$\\
        $\bar\theta^{desc}(x)$&Mean Description-guided image features&$\mathbb{R}^{d}$\\
        $\hat\theta(x)$&Fused image features&$\mathbb{R}^{d}$\\
        $\theta_{p}(x)$&Prompted Global image feature & $\mathbb{R}^{d}$\\
        $\theta^{l}_{p}(x)$&Prompted Local image feature & $\mathbb{R}^{M\times d}$\\
        $\theta^{desc}_{p}(x)$&Prompted Description-guided image features&$\mathbb{R}^{N\times d}$\\
        $\mathbf{r}$&Description relevance score for an image & $\mathbb{R}^{N}$\\
        $\alpha$&average specificity for all descriptions&$\mathbb{R}$\\
        \bottomrule
    \end{tabular}
    
    \vspace{5pt}
    \caption{Notations used in this paper and their descriptions.}
    \label{notationstable}
\end{table}

\section{SAP: Algorithm}
\label{app sec algorithm}

Algorithm~\ref{proposed algo} outlines the \methodname{} methodology. The algorithm is summarized as follows: In a given dataset, descriptions for each class are acquired by querying the LLM (L1 - L4). Class description features are then derived by passing the descriptions through $\phi$ (L5). Unprompted and prompted image features are obtained by processing images through $\theta$ (L7-L8). The description-guided image features are obtained via a parameter-free cross-attention between local features and description features (L9). The local image features are a weighted average of the description-guided features based on the relevance of each description to the image (L10 - L11). Finally, the mean description-guided image features and global image features are fused to create the fusion image feature (L12). Unprompted and prompted description-guided text features are obtained by passing the description-guided text templates through $\phi$ (L13-L14).  $L_{ce}$, $L_{steer}^{v}$, and $L_{steer}^{t}$ loss functions are employed to train the prompts.

\vspace{10pt}
\begin{algorithm}
    \caption{\methodname{} Algorithm}
        \label{proposed algo}
    \footnotesize
    \KwIn{Dataset $D = \{\mathbf{x}_i, y_i\}_{i=1}^{B}$; Classification label space: $\mathcal{Y}$; Vision and Language encoders: ($\theta$, $\phi$); LLM: ChatGPT-3.5 model; Hyperparameters: coefficients $\lambda_1, \lambda_2$, scaling parameter $s$, learning rate $\delta$; Learnable Prompts: $\rho=\{\rho_t, \rho_v\}$}
    \KwOut{Trained parameters $\hat{\rho}$}

    \tcp{Get descriptions for each class by querying LLM}
    \For{each $y \in \mathcal{Y}$}{
        $A_{y} = \text{LLM}(\text{Visual features for distinguishing } y \text{ in a photo?})$
    }
    $A = \bigcup\limits_{y \in \mathcal{Y}} A_{y}$\\
    $\phi(A)$ \tcp*{Get class description features}

    \For{each epoch}{
        \tcp{Get unprompted and prompted image features for every image $\mathbf{x}$ in the batch}
        $\theta(\mathbf{x})$, \_ = $\theta(\mathbf{x})$\\
        $\theta_p(\mathbf{x})$, $\theta_p^l(\mathbf{x})$ = $\theta(\mathbf{x}; \rho_v)$\\
        
        \tcp{Get description-guided image features using parameter-free cross-attention}
        $\theta^{desc}(\mathbf{x}) = \text{Cross\_Attention}(Q=\phi(A), K=\theta^l(\mathbf{x}), V=\theta^l(\mathbf{x}))$\\

        \tcp{Get mean description-guided image feature using relevance score}
        $\mathbf{r} = \text{softmax}(\phi(A) \cdot \theta(\mathbf{x}))$\\
        $\bar{\theta}^{desc}(\mathbf{x}) = \theta^{desc}(\mathbf{x})^{\intercal} \cdot \mathbf{r}$\\

        \tcp{Get fused image feature by fusing global and local feature using description specificity ($\alpha$)}
        $\hat{\theta}(\mathbf{x}) = (1 - \alpha) \cdot \theta(\mathbf{x}) + \alpha \cdot \bar{\theta}^{desc}(\mathbf{x})$\\

        \tcp{Get unprompted and prompted description guided text features for every class $y$}
        $\phi(y, A_y) = \phi(y, A_y)$\\
        $\phi_p(y, A_y) = \phi(y, A_y; \rho_t)$\\

        \tcp{Similarity between an image and a class is the aggregate of similarities over pertinent descriptions of a class}
        $\xi(\hat{\theta}_p(\mathbf{x}), \phi_p(y; A_y)) = \frac{1}{|A_y|} \sum\limits_{a \in A_y} \text{sim}(\hat{\theta}_p(\mathbf{x}), \phi_p(y; a))$\\

        $L_{ce}(\rho) = -\frac{1}{B} \sum\limits_{i=1}^B \log \frac{\exp(\xi(\hat{\theta}_p(\mathbf{x_i}), \phi_p(y_i; A_{y_i}))/\tau)}{\sum\limits_{y \in \mathcal{Y}} \exp(\xi(\hat{\theta}_p(\mathbf{x_i}), \phi_p(y; A_y))/\tau)}$\\

        \tcp{Compute Steering Losses}
        $L_{steer}^v(\rho) = \frac{1}{B} \sum\limits_{i=1}^B \lVert \theta_p(\mathbf{x_i}) - \theta(\mathbf{x_i}) \rVert_1$\\
        $L_{steer}^t(\rho) = \frac{1}{|\mathcal{Y}|} \sum\limits_{y \in \mathcal{Y}} \lVert \phi_p(y; A_y) - \phi(y; A_y) \rVert_1$\\

        \tcp{Perform gradient descent on the total loss}
        $\mathcal{L}(\rho) = L_{ce}(\rho) + \lambda_1 L_{steer}^v(\rho) + \lambda_2 L_{steer}^t(\rho)$\\
        $\hat{\rho} = \rho - \delta \nabla \mathcal{L}(\rho)$\\
    }

    \Return $\hat{\rho}$
\end{algorithm}

\section{Implementation Details}
\label{app sec experimental setup}

\noindent\textbf{Training Details.} We use the ViT-B/16~\citep{dosovitskiy2021an}-based CLIP model as our backbone. For the GZS and B2N benchmarks, we fine-tune the model on $K=16$ shot training data from the base classes. Prompts are learned in the first three layers for the Cross-dataset benchmark and the first nine layers for the remaining two benchmarks. We introduce a $d$-dimensional bias as the sole additional parameter compared to ~\citep{promptsrc}. The text prompts in the initial layer are initialized with the word embeddings of \texttt{`a photo of a'}, and the rest are randomly initialized from a normal distribution, similar to~\citep{promptsrc}. Our models are trained on a single Tesla V100 GPU with Nvidia driver version 470.199.02. We train for $20$ epochs, with a batch size of 4 images, $\lambda_1=10$ and $\lambda_2=25$. The hyperparameter setup is common across all datasets. We use the SGD optimizer with a momentum of $0.9$, a learning rate of $0.0025$, and weight decay $5e-4$. A cosine learning rate scheduler is applied with a warmup epoch of $1$. Image pre-processing involves random crops, random horizontal and vertical flips, and normalization using mean values of $[0.48, 0.46, 0.41]$ and standard deviation values of $[0.27, 0.26, 0.27]$. All baselines utilize publicly available codes and models. All results are averages over three seeds. We use PyTorch 1.12, CUDA 11.3, and build on the Dassl code repository: \href{https://github.com/KaiyangZhou/Dassl.pytorch}{https://github.com/KaiyangZhou/Dassl.pytorch}. We will open-source our code on acceptance.
\section{Expanded Tables and Additional Results}
\label{app sec additional results}
\noindent \textbf{Using Random Text in place of Class Descriptions.} To study the usefulness of valid descriptions, we replace the descriptions for each class by randomly generated texting in Tab.~\ref{tab: random descriptions}. Examples of random descriptions are ``Raindrops pattered softly against the roof'', ``A solitary figure walked down the empty street''. We observe that descriptions matter for unusual datasets having texture-based images, satellite images, aircraft images and action recognition images. The average HM using random text across 11 datasets on B2N benchmark is \textbf{78.27\%}, while \methodname{} reports an average HM of \textbf{80.94\%}. A drop of \textbf{2.67\%} is noted.

\vspace{4pt}
\begin{table}
    \centering
    \scalebox{0.85}{
    \begin{tabular}{lccccccccccc>{\columncolor[gray]{0.8}}c}
    \toprule
         & \rotatebox{90}{UCF101}&\rotatebox{90}{EuroSAT}&\rotatebox{90}{DTD}&\rotatebox{90}{OxfordPets}&\rotatebox{90}{StanfordCars}&\rotatebox{90}{Flowers102}&\rotatebox{90}{Food101}&\rotatebox{90}{FGVCAircraft}&\rotatebox{90}{SUN397}&\rotatebox{90}{Caltech101}&\rotatebox{90}{ImageNet}&\rotatebox{90}{Average}\\
    \midrule

    Base&\large 86.27&\large 95.83&\large 83.1&\large 95.07&\large 78.2&\large 97.5&\large 90.13&\large 41.37&\large 81.87&\large 98.07&\large 76.7&\large 84.01\\

    Novel&\large 76.37&\large 69.23&\large 54.1&\large 95.33&\large 72.33&\large 75.53&\large 89.9&\large 34.8&\large 76.63&\large 94.1&\large 67.7&\large 73.27\\
    
    \midrule
    
    HM&\large 81.02&\large 80.39&\large 65.54&\large 95.2&\large 75.15&\large 85.12&\large 90.01&\large 37.8&\large 79.16&\large 96.04&\large 72.17&\large 78.27\\
    \bottomrule
    \end{tabular}
    }
    \vspace{4pt}
    \caption{B2N benchmark results using random text in place of class descriptions. The results show that using irrelevant descriptions hurts model performance.}

    \label{tab: random descriptions}
\end{table}

\begin{figure}
    \centering
    \includegraphics[width=0.75\textwidth]{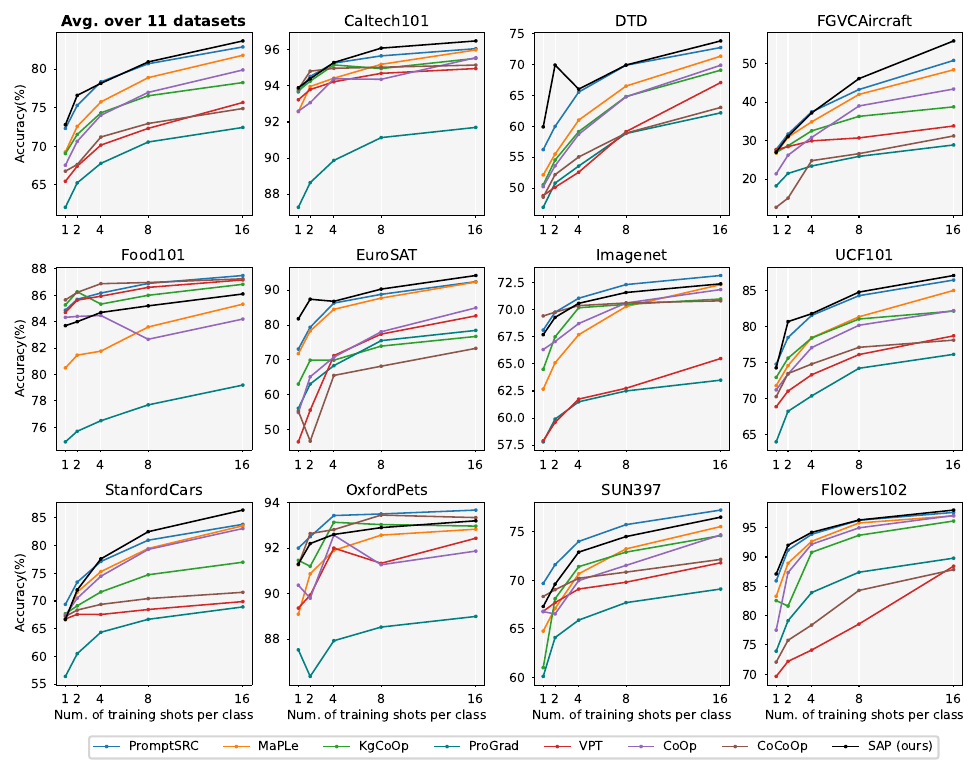}
    \caption{Performance of \methodname{} in the few-shot setting. Our method achieves competitive performance compared to all baselines on average across 11 datasets.}
    \label{fewshotexpts}
\end{figure}

\noindent \textbf{Few-shot Setting.} Our main objective is to train prompts that can generalize effectively to novel classes and datasets.
As such, we present results primarily on settings that test generalizability, such as the GZS benchmark, Base-to-Novel benchmark, and the Out-of-Vocabulary benchmark. For completeness, we present results in a few-shot classification setting, where limited training samples are provided for all classes. Note that there are no novel classes in this setting. We showcase outcomes for $K=1, 2, 4, 8, \text{and } 16$ shots. As shown in Fig.~\ref{fewshotexpts}, on average, across 11 datasets, we perform competitively against the best baseline PSRC.

\begin{wraptable}[7]{r}{0.5\textwidth}
    \centering
    \scalebox{0.750}{
    \begin{tabular}{lccccc>{\columncolor[gray]{0.8}}c}
    \toprule
         &Source&\multicolumn{4}{c}{Target} \\
    \cmidrule(lr){2-2}\cmidrule(lr){3-6}
        &ImageNet&-V2&-A&-S&-R&Avg\\
    \midrule
    MaPLe&77.10&71.00&53.70&50.00&77.70&63.10\\
    PSRC&76.30&71.00&54.10&50.00&77.80&63.22\\
    \midrule
    \methodname{}&76.40&71.10&55.70&49.80&77.50&\textbf{63.52}\\
    \bottomrule
    \end{tabular}
    }
    \vspace{-5pt}
    \caption{DG benchmark. \methodname{} outperforms baselines on avg. }

    \label{domaingeneralization}
\end{wraptable} 

\noindent\textbf{Domain Generalization.} We show results on Domain Generalization in Tab.~\ref{domaingeneralization}. We train on $K=16$ shot training data from base classes of source dataset ImageNet and evaluation on ImageNetV2, ImageNet-A, ImageNet-Setch, and ImageNet-R target datasets.  \methodname{} outperforms two strong baselines PSRC and MaPLe.

\noindent\textbf{ResNet-50 Backbone as Image Encoder.} Here we show the GZS and B2N performance of \methodname{} using the ResNet-50 CLIP model as a backbone. We compare against five baselines which also use the ResNet-50 backbone and present our results in Tab.~\ref{generalizedresnet}. For all methods including ours, we train the models without tuning any hyperparameters such as prompt-depth, regularization weight, learning rate etc. and use the same values as those of ViT-B/16 CLIP backbone. We observe that PSRC performs particularly poorly with a ResNet backbone. Although we use similar hyperparameters as PSRC, \methodname{} shows good results, indicating that class descriptions help greatly in this setting. We show a gain of $\textbf{+0.98\%}$ on average gHM for GZS, and $\textbf{+2.32\%}$ on average HM in the B2N setting.
\vspace{4pt}
\begin{table}[H]
    \centering
    \footnotesize
    \scalebox{1}{
    \begin{tabular}{cc|ccccc|c}
    \toprule
         \textbf{Dataset}& &\textbf{CLIP}&\textbf{CoOp}&\textbf{KgCoOp}&\textbf{ProGrad}&\textbf{PSRC}&\textbf{\methodname{} (Ours)}\\
     \midrule
     \multicolumn{8}{c}{\textbf{Generalized Zero-Shot Learning Benchmark}}\\
     \midrule
         \multirow{3}{4em}{\textbf{Average on 11 datasets}} & gBase &57.01 & 68.65 & 69.25  & \underline{69.89} & 47.41 & \textbf{71.52 (\textcolor{darkgreen}{+1.63})}\\
         & gNovel &\textbf{60.73} & 50.35 & 59.08 & 52.26 & 29.16 & \underline{59.13}\textbf{ (\textcolor{red}{-1.60})}\\
         & gHM &58.81 & 58.1 & \underline{63.76} & 59.81 & 36.12 & \textbf{64.74 (\textcolor{darkgreen}{+0.98})}\\
         \midrule
         \multicolumn{8}{c}{\textbf{Base-to-Novel Generalization Benchmark}}\\
    \midrule
         \multirow{3}{4em}{\textbf{Average on 11 datasets}} & Base &65.27&77.24&75.51&\underline{77.98}&55.13&\textbf{78.49 (\textcolor{darkgreen}{+0.51})}\\
         & Novel &68.14&57.40&\underline{67.53}&63.41& 38.72&\textbf{69.32 (\textcolor{darkgreen}{+1.79})}\\
         & HM &66.68&65.86&\underline{71.30}&69.94& 45.49&\textbf{73.62 (\textcolor{darkgreen}{+2.32})}\\
             \bottomrule
    \end{tabular}
    }
    \vspace{3pt}
    \caption{Results on GZS and B2N settings using a ResNet-50 backbone. On average, \methodname{} outperforms all the baselines.}
    \label{generalizedresnet}
\end{table}

\begin{wraptable}[4]{r}{0.4\textwidth}
    \centering
    \scalebox{0.7}{
    \begin{tabular}{ccccccc}
    \toprule
         Depth&1&3&5&7&9&11 \\
         \midrule
         HM&76.84&79.35&79.25&80.85&\textbf{81.76}&80.68\\
        \bottomrule
    \end{tabular}
    }
    \vspace{-5pt}
    \caption{Prompt depth analysis}
    \label{tab:prompt depth}
\end{wraptable}

\noindent \textbf{Prompt Depth.}
Tab.~\ref{tab:prompt depth} shows the average HM for the B2N benchmark across nine datasets, excluding SUN397 and ImageNet. As seen from the table, adding prompts till depth 9 for image and text encoders is ideal for \methodname{} performance and is used for B2N, GZS and OVC benchmarks.

\begin{figure}[H]
    \centering
    \includegraphics[width=0.8\textwidth]{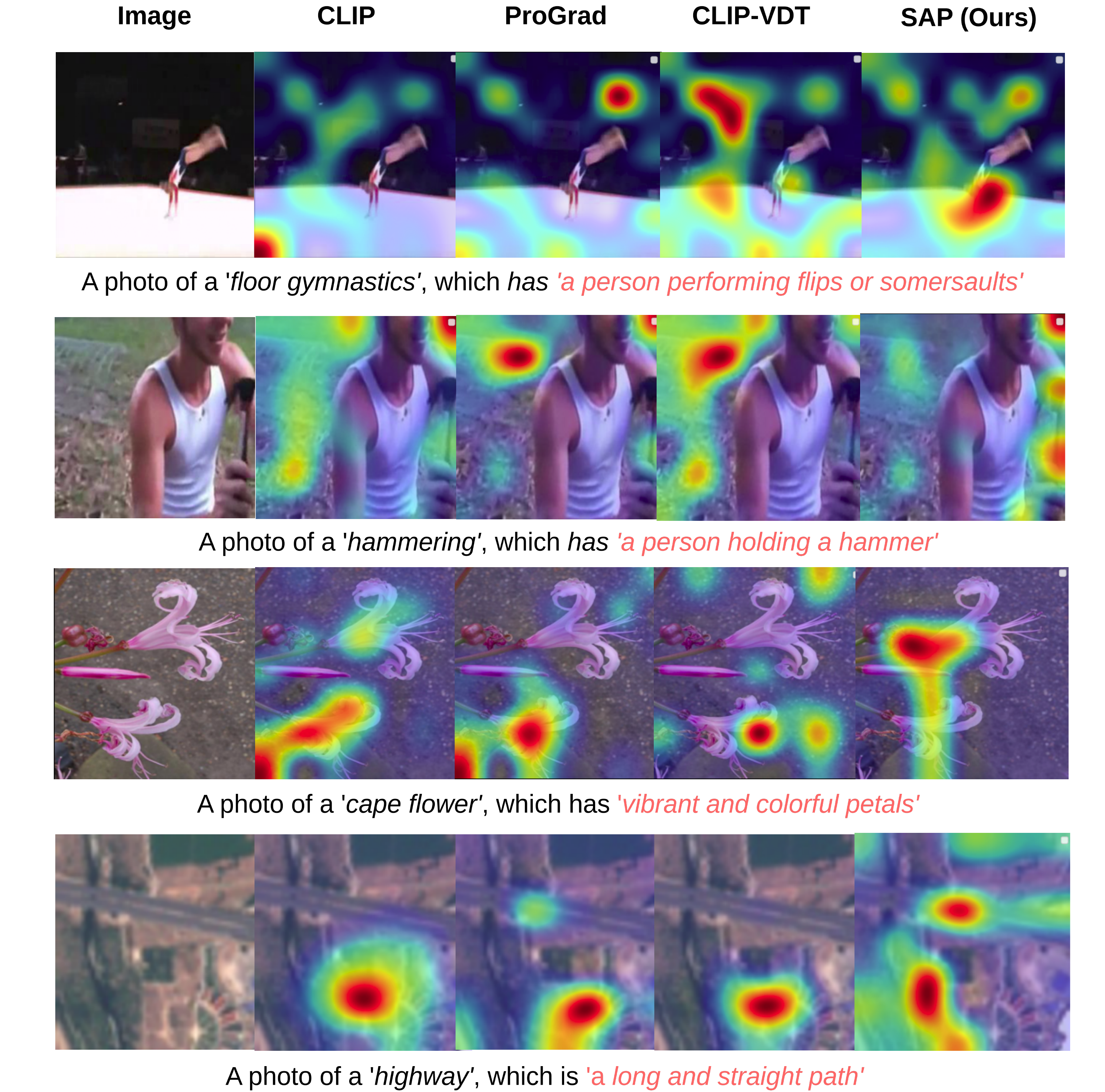}
    \caption{Figure displays GradCAM visualizations that highlight the regions of highest activation relevant to specific text phrases. These visualizations use a ResNet-50 backbone as the image encoder for all baselines, including ours. \methodname{} localizes better than the existing baselines.}
    \label{fig:additional_saliency}
\end{figure}

\noindent\textbf{Additional Class Activation Maps (CAMs).} We show additional CAMs for the ResNet-50\citep{resnet} backbone encoder to visualize image regions that most correlate to a given description.
Fig.~\ref{fig:additional_saliency} shows the GradCAM~\citep{gradcam} visualizations for base classes \textit{``Floor gymnastics''}, \textit{``Hammering''}, \textit{``Cape Flower''} and \textit{``Highway''}. \methodname{} effectively localizes the text semantics in the image compared to baselines.

\noindent\textbf{Expanded Dataset-wise Tables.} We present the elaborate tables dataset-wise for the Generalized Zero-Shot setting in Tab.~\ref{gzs table} and  Base-to-Novel generalization setting in Tab.~\ref{basetonewappendix}. \methodname{} outperforms the best-performing baseline, PSRC, in 7 of the 11 considered datasets. We perform very well in challenging datasets such as EuroSAT, DTD, and UCF-101. We present dataset-wise results for the Out-of-Vocabulary benchmark in Tab.~\ref{classificationwoclassnameappendix}. Tab.~\ref{crossdatasetappendix} has the dataset-wise results for the Cross-Dataset generalizatin benchmark. In Tab.~\ref{generalizedresnet} we show average results on the GZS benchmark and the Base-to-Novel benchmark for the ResNet-50 backbone Image Encoder. We also present detailed, dataset-wise results for the same in Tab.~\ref{gzsresnetappendix}.

\begin{table}
    \centering
    \footnotesize
    \scalebox{0.7}{
    \begin{tabular}{cc|ccccccccc|c}
    \toprule
         \textbf{Dataset}& &\textbf{CLIP}&\textbf{CoOp}&\textbf{VPT}&\textbf{CoCoOp}&\textbf{MaPLe}&\textbf{KgCoOp}&\textbf{ProGrad}&\textbf{PSRC}&\textbf{CLIP-VDT}&\textbf{\methodname{}}\\
        &&(ICML '21)&(IJCV '22)&(ECCV '22)&(CVPR '22)&(CVPR '23)&(CVPR '23)&(ICCV '23)&(ICCV '23)&(ICCVW '23)&(Ours)\\
     \midrule
     \rowcolor{Gray}\textbf{Average}& gBase&60.81&75.19&73.48&73.13&75.47&76.86&70.15&\underline{78.81}&63.75&\textbf{79.47 (\textcolor{darkgreen}{+0.66})}\\
     \rowcolor{Gray}\textbf{on 11}&gNovel&63.21&60.39&66.62&65.23&67.09&62.12&55.07&\underline{68.13}&63.89&\textbf{69.75 (\textcolor{darkgreen}{+1.62})}\\
     \rowcolor{Gray}\textbf{datasets}&gHM&61.99&66.99&69.89&68.96&71.04&68.71&61.70&\underline{73.08}&63.82&\textbf{74.29 (\textcolor{darkgreen}{+1.21})}\\
      \midrule
      \multirow{3}{4em}{UCF101} & gBase&62.70&80.26&75.76&76.56&76.90&78.96&74.63&\textbf{82.67}&66.19&\underline{82.23}\\
     &gNovel&64.40&\textbf{84.76}&67.73&64.76&70.40&62.33&51.36&71.40&67.00&\underline{76.40}\\
     &gHM&63.53&\textbf{82.45}&71.52&70.17&73.51&69.67&60.85&76.62&66.59&\underline{79.21}\\
      \midrule
     \multirow{3}{4em}{EuroSAT} & gBase&51.40&69.26&\underline{88.22}&70.86&84.06&82.02&76.26&86.60&55.09&\textbf{94.37}\\
     &gNovel&38.90&36.26&53.36&41.03&43.90&31.26&23.43&\underline{54.16}&50.79&\textbf{58.53}\\
     &gHM&44.28&47.60&66.50&51.97&57.68&45.28&35.85&\underline{66.65}&52.85&\textbf{72.25}\\
     \midrule
     \multirow{3}{4em}{DTD} & gBase&42.70&65.36&58.92&60.29&63.00&66.42&57.19&\textbf{68.73}&55.79&\underline{66.47}\\
     &gNovel&45.79&34.30&44.26&46.09&47.49&39.73&33.36&\underline{47.53}&51.00&\textbf{54.27}\\
     &gHM&44.19&44.99&50.55&52.25&54.16&49.72&42.14&\underline{56.20}&53.28&\textbf{59.75}\\
      \midrule
     \multirow{3}{4em}{Oxford Pets} & gBase&84.80&89.56&89.06&91.12&91.69&\underline{91.99}&88.36&\textbf{93.00}&83.80&91.97\\
     &gNovel&90.19&90.46&93.23&92.50&\textbf{93.93}&\underline{92.69}&87.76&91.00&90.40&92.30\\
     &gHM&87.41&90.01&91.10&91.81&\textbf{92.80}&\underline{92.34}&88.06&91.99&86.97&92.13\\
      \midrule
     \multirow{3}{4em}{Stanford Cars} & gBase&56.00&74.43&65.13&67.29&69.33&72.56&64.46&\underline{74.77}&59.50&\textbf{76.40}\\
     &gNovel&64.19&57.16&\underline{70.56}&68.82&69.86&66.56&55.66&\textbf{71.23}&61.59&69.33\\
     &gHM&59.81&64.67&67.74&68.05&69.61&69.43&59.74&\textbf{72.96}&60.52&\underline{72.69}\\
      \midrule
     \multirow{3}{4em}{Flowers102} & gBase&62.09&93.40&83.12&87.36&91.19&92.80&84.86&\underline{95.00}&69.90&\textbf{95.69}\\
     &gNovel&69.80&56.92&65.56&65.53&68.29&65.76&62.39&\underline{71.00}&77.00&\textbf{71.13}\\
     &gHM&65.71&70.74&73.31&74.89&78.10&76.97&71.92&\underline{81.27}&73.20&\textbf{81.60}\\
      \midrule
     \multirow{3}{4em}{Food101} & gBase&79.90&83.59&85.96&86.15&\underline{86.76}&85.76&78.46&\textbf{87.07}&75.90&86.43\\
     &gNovel&80.90&76.82&84.99&\underline{86.50}&\textbf{87.20}&83.72&76.23&85.90&77.69&86.09\\
     &gHM&80.39&80.07&85.49&86.33&\textbf{86.98}&84.73&77.33&\underline{86.48}&76.78&86.26\\
      \midrule
     \multirow{3}{4em}{FGVC Aircraft} & gBase&14.50&29.92&25.12&25.90&25.90&32.69&23.93&\underline{34.90}&16.10&\textbf{35.00}\\
     &gNovel&23.79&22.83&28.03&26.36&\underline{28.53}&22.06&15.63&28.40&18.60&\textbf{30.23}\\
     &gHM&18.01&25.90&26.50&26.13&27.15&26.35&18.93&\underline{31.32}&17.59&\textbf{32.44}\\
      \midrule
     \multirow{3}{4em}{SUN397} & gBase&60.50&72.56&69.40&71.19&72.76&73.36&67.69&\textbf{75.63}&63.09&\underline{75.40}\\
     &gNovel&63.70&56.52&67.50&67.26&\underline{68.93}&61.75&57.00&68.70&66.00&\textbf{69.80}\\
     &gHM&62.05&63.55&68.44&69.17&70.79&67.06&61.89&\underline{72.00}&64.51&\textbf{72.30}\\
     \midrule
     \multirow{3}{4em}{Caltech101} & gBase&91.40&95.92&95.66&95.09&95.83&95.89&91.53&96.20&93.59&\textbf{96.30}\\
     &gNovel&91.69&85.09&\underline{92.26}&90.93&92.03&92.06&85.26&91.73&86.19&\textbf{92.82}\\
    &gHM&91.54&90.19&93.94&92.97&93.89&\underline{93.94}&88.29&\underline{93.91}&89.73&\textbf{94.53}\\
     \midrule
    \multirow{3}{4em}{Imagenet} & gBase&63.00&72.80&71.9&72.59&72.80&\underline{73.00}&64.19&72.30&61.79&\textbf{73.97}\\
     &gNovel&62.00&63.20&65.40&67.80&67.40&65.40&57.70&\textbf{68.40}&56.59&66.66\\
     &gHM&62.49&67.66&68.50&70.11&70.00&68.99&60.77&\textbf{70.30}&59.07&\underline{70.13}\\
     \bottomrule
     \end{tabular}
     }
    \vspace{3pt}
     \caption{Accuracy comparison on the GZS benchmark. gNovel \& gBase indicate the accuracy of the novel classes and base classes respectively under the joint classification label space. gHM is the harmonic mean of gBase and gNovel. The best numbers are in bold, and the second best are underlined. As reported in the first row, \methodname{} outperforms all baselines on average gBase (by $+0.66\%$), gNovel (by $+1.62\%$), and gHM (by $1.21\%$) computed across all datasets. We indicate the margin of improvement over the corresponding best-performing baseline for each metric in green. }
     \label{gzs table}
\end{table}

\begin{table*}
    \centering
    \footnotesize
    \scalebox{0.9}{
    \begin{tabular}{cc|cccccccc|c}
    \toprule
         \textbf{Dataset}& &\textbf{CLIP}&\textbf{CoOp}&\textbf{VPT}&\textbf{CoCoOp}&\textbf{MaPLe}&\textbf{KgCoOp}&\textbf{ProGrad}&\textbf{PSRC}&\textbf{\methodname{}}\\
     \midrule
         \rowcolor{Gray}{\textbf{Average}} & Base&33.28&36.97&40.28&40.12&\underline{41.56}&37.95&34.00&40.40&\textbf{43.31 (\textcolor{darkgreen}{+1.75})}  \\
         \rowcolor{Gray}{\textbf{on 11}} & Novel &38.55&\underline{43.90}&43.72&40.80&43.30&40.69&35.01&43.78&\textbf{45.66 (\textcolor{darkgreen}{+1.76})} \\
        \rowcolor{Gray}{\textbf{datasets}}  & HM&35.72&40.14&41.93&40.46&\underline{42.41}&39.27&34.50&42.02&\textbf{44.46 (\textcolor{darkgreen}{+2.04})}\\
    \midrule
         \multirow{3}{4em}{UCF101} & Base &56.60&61.20&61.20&61.70&\underline{64.20}&62.00&59.70&63.10&\textbf{64.70}\\
         & Novel &62.20&66.80&63.20&\textbf{70.70}&\underline{70.40}&68.80&63.50&69.40&69.10\\
         & HM&59.27&63.88&62.18&65.89&\textbf{67.16}&65.22&61.54&66.10&\underline{66.83}\\
         \midrule
        \multirow{3}{4em}{EuroSAT} & Base &39.90&47.10&76.50&62.90&\underline{84.30}&59.70&47.60&71.4&\textbf{88.70}\\
         & Novel &71.10&78.70&\textbf{83.20}&49.00&58.30&57.60&45.80&\underline{82.10}&80.90\\
         & HM&51.12&58.93&\underline{79.71}&55.09&68.93&58.63&46.68&76.38&\textbf{84.62}\\
         \midrule
        \multirow{3}{4em}{DTD} & Base &40.20&40.90&\underline{47.20}&44.20&44.90&41.90&39.20&42.70&\textbf{52.40}\\
         & Novel &42.40&44.10&44.30&\underline{47.10}&42.90&44.40&40.20&44.00&\textbf{49.00}\\
         & HM&41.27&42.44&\underline{45.70}&45.60&43.88&43.11&39.69&43.34&\textbf{50.64}\\
    \midrule
    \multirow{3}{4em}{Oxford Pets} & Base &24.50&32.00&22.30&\textbf{34.20}&\underline{32.80}&25.40&23.10&27.40&23.60\\
         & Novel &35.20&40.80&40.70&\underline{44.10}&\textbf{46.40}&39.70&36.00&41.60&\underline{44.10}\\
         & HM&28.89&35.87&28.81&\textbf{38.52}&\underline{38.43}&30.98&28.14&33.04&30.75\\
    \midrule
    \multirow{3}{4em}{Stanford Cars} & Base &13.50&15.60&17.60&16.30&10.30&12.50&10.00&\underline{21.00}&\textbf{22.50}\\
         & Novel &15.90&20.70&18.90&11.70&\textbf{25.80}&15.30&8.50&20.40&\underline{23.40}\\
         & HM&14.60&17.79&18.23&13.62&14.72&13.76&9.19&\underline{20.70}&\textbf{22.94}\\
    \midrule
    \multirow{3}{4em}{Flowers102} & Base &7.40&14.10&12.40&17.70&18.30&12.00&16.40&\underline{18.80}&\textbf{19.60}\\
         & Novel &9.30&20.40&18.40&17.60&\underline{23.20}&12.30&13.80&19.30&\textbf{26.00}\\
         & HM&8.24&16.67&14.82&17.65&\underline{20.46}&12.15&14.99&19.05&\textbf{22.35}\\
    \midrule
    \multirow{3}{4em}{Food101} & Base &35.10&42.70&\textbf{44.00}&\underline{43.40}&35.50&47.10&42.10&41.20&42.20\\
         & Novel &33.80&\textbf{45.40}&\underline{44.80}&44.40&38.90&44.60&41.80&40.50&44.20\\
         & HM&34.44&\underline{44.01}&44.40&43.89&37.12&\textbf{45.82}&41.95&40.85&43.18\\
    \midrule
    \multirow{3}{4em}{FGVC Aircraft} & Base &6.10&\underline{9.50}&8.00&7.00&\textbf{13.40}&6.80&5.20&8.30&9.40\\
         & Novel &7.90&\textbf{15.80}&12.80&8.30&\underline{15.50}&10.70&8.20&12.30&12.30\\
         & HM&6.88&\underline{11.87}&9.85&7.59&\textbf{14.37}&8.32&6.36&9.91&10.66\\
    \midrule
    \multirow{3}{4em}{SUN397} & Base &46.60&49.20&50.50&\underline{51.30}&50.20&50.10&40.10&50.00&\textbf{51.40}\\
         & Novel &48.30&50.00&51.40&\underline{52.50}&52.20&\textbf{53.20}&42.90&51.40&51.40\\
         & HM&47.43&49.60&50.95&\textbf{51.89}&51.18&\underline{51.60}&41.45&50.69&51.40\\
    \midrule
    \multirow{3}{4em}{Caltech101} & Base &77.80&76.00&\textbf{83.00}&\textbf{83.00}&82.30&80.80&72.30&81.10&81.70\\
         & Novel &74.80&74.30&\underline{75.90}&75.80&75.50&\textbf{76.20}&63.20&75.10&75.20\\
         & HM&76.27&75.14&\textbf{79.29}&\underline{79.24}&78.75&78.43&67.44&77.98&78.32\\
    \midrule
    \multirow{3}{4em}{ImageNet} & Base &18.40&18.40&\underline{20.40}&19.70&\textbf{21.00}&19.20&18.30&19.4&20.30\\
         & Novel &23.20&26.00&\underline{27.40}&\textbf{27.60}&27.30&24.80&21.30&25.50&26.70\\
         & HM&20.52&21.55&\underline{23.39}&22.99&\textbf{23.74}&21.64&19.69&22.04&23.06\\
    \bottomrule
    \end{tabular}
    }
    \vspace{3pt}
    \caption{Accuracy comparison in the Out-of-Vocabulary setting. We show average Base, Novel, and HM accuracies over all 11 datasets. During evaluation, descriptions of each class are provided instead of the class name, and visual recognition is conducted based on these descriptions. \methodname{} outperforms baselines by average Base (by $+1.75\%$), Novel (by $+1.76\%$) and HM (by $+2.04\%$) computed over all datasets.}
    \label{classificationwoclassnameappendix}
\end{table*}

\begin{table}
    \centering
    \footnotesize
    \scalebox{0.58}{
    \begin{tabular}{cc|cccccccccccc|c}
    \toprule
         \textbf{Dataset}& &\textbf{CLIP}&\textbf{CoOp}&\textbf{VPT}&\textbf{CoCoOp}&\textbf{ProDA}&\textbf{MaPLe}&\textbf{KgCoOp}&\textbf{ProGrad}&\textbf{PSRC}&\textbf{L.Prompt}&\textbf{CLIP-VDT}&\textbf{KAPT}&\textbf{\methodname{}}\\
     \midrule
         \rowcolor{Gray}{\textbf{Average}} & Base &69.34&82.69&80.81&80.47&81.56&82.28&80.73&82.48&84.26&\underline{84.47}&82.48&81.10&\textbf{84.68 (\textcolor{darkgreen}{+0.21})} \\
         \rowcolor{Gray}{\textbf{on 11}}& Novel &74.22&63.22&70.36&71.69&72.30&75.14&73.60&70.75&\underline{76.10}&74.24&74.50&72.24&\textbf{77.51 (\textcolor{darkgreen}{+1.41})} \\
         \rowcolor{Gray}{\textbf{datasets}}& HM &71.70&71.66&70.36&75.83&76.65&78.55&77.00&76.16&\underline{79.97}&79.03&78.28&76.41&\textbf{80.94 (\textcolor{darkgreen}{+0.97})} \\
    \midrule
         \multirow{3}{4em}{UCF101} & Base &70.53&84.69&82.67&82.33&85.23&83.00&82.89&84.33&\textbf{87.10}&86.19&84.10&80.83&\underline{86.60}\\
         & Novel &77.50&56.05&74.54&77.64&78.04&\underline{80.77}&76.67&76.94&78.80&73.07&76.40&67.10&\textbf{83.90}\\
         & HM&73.85&67.46&78.39&77.64&78.04&80.77&79.65&79.35&\underline{82.74}&79.09&80.07&73.33&\textbf{85.23}\\
         \midrule
        \multirow{3}{4em}{EuroSAT} & Base &56.48&92.19&93.01&87.49&83.90&\underline{94.07}&85.64&90.11&92.90&93.67&88.50&84.80&\textbf{96.10}\\
         & Novel &64.05&54.74&54.89&60.04&66.00&73.23&64.34&60.89&\underline{73.90}&69.44&70.50&67.57&\textbf{81.13}\\
         & HM&60.03&68.69&69.04&71.21&73.88&\underline{82.35}&73.48&72.67&82.32&79.75&78.48&75.21&\textbf{87.98} \\
         \midrule
        \multirow{3}{4em}{DTD} & Base &53.24&79.44&79.15&77.01&80.67&80.36&77.55&77.35&\underline{83.37}&82.87&81.80&75.97&\textbf{84.27}\\
         & Novel &59.90&41.18&50.76&56.00&56.48&59.18&54.99&52.35&\underline{62.97}&60.14&62.30&58.30&\textbf{67.03} \\
         & HM&56.37&54.24&61.85&64.85&66.44&68.16&64.35&62.45&\underline{71.75}&69.70&70.73&65.97&\textbf{74.67} \\
    \midrule
    \multirow{3}{4em}{Oxford Pets} & Base &91.17&93.67&94.81&95.20&\underline{95.43}&\underline{95.43}&94.65&95.07&95.33&\textbf{96.07}&94.40&93.13&95.27\\
         & Novel &97.26&95.29&96.00&97.69&\textbf{97.83}&\underline{97.76}&\underline{97.76}&97.63&97.30&96.31&97.70&96.53&96.90 \\
         & HM&94.12&94.47&95.40&96.43&\textbf{96.62}&\underline{96.58}&96.18&96.33&96.30&96.18&95.68&94.80&96.08 \\
    \midrule
    \multirow{3}{4em}{Stanford Cars} & Base &63.37&78.12&72.46&70.49&74.70&72.94&71.76&77.68&78.27&\underline{78.36}&76.80&69.47&\textbf{79.70}\\
         & Novel &74.89&60.40&73.38&73.59&71.20&74.00&\textbf{75.04}&68.63&\underline{74.97}&72.39&72.90&66.20&73.47\\
         & HM&68.65&68.13&72.92&72.01&72.91&73.47&73.36&72.88&\textbf{76.58}&75.26&74.80&67.79&\underline{76.46}\\
    \midrule
    \multirow{3}{4em}{Flowers102} & Base &72.08&97.60&95.39&94.87&97.70&95.92&95.00&95.54&\underline{98.07}&\textbf{99.05}&97.40&95.00&97.83\\
         & Novel &\textbf{77.80}&59.67&73.87&71.75&68.68&72.46&74.73&71.87&\underline{76.50}&76.52&75.30&71.20&\underline{76.50}\\
         & HM&74.83&74.06&83.26&81.71&80.66&82.56&83.65&82.03&85.95&\underline{86.34}&84.94&81.40&\textbf{86.86}\\
    \midrule
    \multirow{3}{4em}{Food101} & Base &90.10&88.33&89.88&90.70&90.30&\underline{90.71}&90.50&90.37&90.67&\textbf{90.82}&90.40&86.13&90.40 \\
         & Novel &91.22&82.26&87.76&91.29&88.57&\textbf{92.05}&\underline{91.70}&89.59&91.53&91.41&91.20&87.06&91.43\\
         & HM&90.66&85.19&88.81&90.99&89.43&\textbf{91.38}&91.09&89.98&91.10&\underline{91.11}&90.80&86.59&90.91 \\
    \midrule
    \multirow{3}{4em}{FGVC Aircraft} & Base &27.19&40.44&33.10&33.41&36.90&37.44&36.21&40.54&42.73&\textbf{45.98}&37.80&29.67&\underline{42.93} \\
         & Novel &36.29&22.30&30.49&23.71&34.13&35.61&33.55&27.57&\underline{37.87}&34.67&33.00&28.73&\textbf{38.87}\\
         & HM&31.09&28.75&31.74&27.74&35.46&36.50&34.83&32.82&\underline{40.15}&39.53&35.24&29.19&\textbf{40.80}\\
    \midrule
    \multirow{3}{4em}{SUN397} & Base &69.36&80.60&79.66&79.74&78.67&80.82&80.29&81.26&\textbf{82.67}&81.20&81.40&79.40&\underline{82.57}\\
         & Novel &75.35&65.89&72.68&76.86&76.93&78.70&76.53&74.17&\underline{78.47}&78.12&76.80&74.33&\textbf{79.20}\\
         & HM&72.23&72.51&79.63&78.27&77.79&79.75&78.36&77.55&\underline{80.52}&79.63&79.03&76.78&\textbf{80.85}\\
    \midrule
    \multirow{3}{4em}{Caltech101} & Base &96.84&98.00&97.86&97.96&\underline{98.27}&97.74&97.72&98.02&98.10&98.19&\textbf{98.30}&97.10&98.23\\
         & Novel &94.00&89.91&93.76&93.81&93.23&94.36&\underline{94.39}&93.89&94.03&93.78&\textbf{95.90}&93.53&94.37\\
         & HM&95.40&93.73&95.77&95.84&95.68&96.02&96.03&95.91&96.02&95.93&\textbf{97.09}&95.28&\underline{96.26} \\
    \midrule
    \multirow{3}{4em}{ImageNet} & Base &72.43&76.47&70.93&75.98&75.40&76.66&75.83&\underline{77.02}&\textbf{77.60}&76.74&76.40&71.10&\textbf{77.60} \\
         & Novel &68.14&67.88&65.90&70.43&70.23&70.54&69.96&66.66&\underline{70.73}&\textbf{70.83}&68.30&65.20&69.83 \\
         & HM&70.22&71.92&73.66&73.10&72.72&73.47&72.78&71.46&\textbf{74.01}&\underline{73.66}&72.12&68.02&73.51\\
    \bottomrule
    \end{tabular}
    }
    \vspace{3pt}
    \caption{Accuracy comparison on Base-to-Novel Generalization benchmark. The best numbers are in bold, and the second best are underlined. \methodname{} outperforms all baselines on average Base (by $+0.21\%$), Novel (by $+1.41\%$) and HM (by $+0.97\%$) computed over all datasets. We indicate the margin of improvement over the corresponding best-performing baseline for each metric in green.}
    \label{basetonewappendix}
\end{table}

\begin{table*}
    \centering
    \footnotesize
    \scalebox{0.62}{
    \begin{tabular}{c>{\columncolor[gray]{0.9}}c|ccccc|c|>{\columncolor[gray]{0.9}}c|ccccc|c}
    \toprule
         &\multicolumn{7}{c|}{\textbf{GZS Benchmark}}&\multicolumn{7}{c}{\textbf{Base-to-Novel Benchmark}}\\
         \midrule
         \textbf{Dataset}& &\textbf{CLIP}&\textbf{CoOp}&\textbf{KgCoOp}&\textbf{Pro-}&\textbf{PSRC}&\textbf{\methodname{}}&&\textbf{CLIP}&\textbf{CoOp}&\textbf{KgCoOp}&\textbf{Pro-}&\textbf{PSRC}&\textbf{\methodname{}}\\
         &&&&&\textbf{Grad}&&&&&&&\textbf{Grad}&&\\
         
     \midrule
         \rowcolor{Gray}{\textbf{Average}} & gBase &57.01&68.65&69.25&\underline{69.89}&47.41&\textbf{71.52 (\textcolor{darkgreen}{+1.63})}&Base&65.27&77.24&75.51&\underline{77.98}&55.13&\textbf{78.49 (\textcolor{darkgreen}{+0.51})}\\
         \rowcolor{Gray}{\textbf{on 11}}& gNovel &\textbf{60.73}&50.35&59.08&52.26&29.16&\underline{59.13} (\textcolor{darkred}{-1.60})&Novel&68.14&57.40&\underline{67.53}&63.41&38.72&\textbf{69.32 (\textcolor{darkgreen}{+1.79})}\\
         \rowcolor{Gray}{\textbf{datasets}}& gHM &58.81&58.10&\underline{63.76}&59.81&36.12&\textbf{64.74 (\textcolor{darkgreen}{+0.98})}&HM&66.68&65.86&\underline{71.30}&69.94&45.49&\textbf{73.62 (\textcolor{darkgreen}{+2.32})}\\
    \midrule
         \multirow{3}{4em}{UCF101} & gBase &61.20&\underline{73.20}&71.05&72.75&51.55&\textbf{74.73}&Base&68.40&79.78&77.16&\textbf{81.04}&59.95&\underline{80.70}\\
         & gNovel&61.79&45.10&56.95&48.05&30.25&\textbf{63.80}&Novel&61.50&48.31&\underline{70.13}&60.07&38.85&\textbf{72.67}\\
         & gHM&61.49&55.81&63.22&57.87&38.13&68.33&HM&64.77&60.18&\underline{73.48}&69.00&47.15&\textbf{76.47}\\
         \midrule
        \multirow{3}{4em}{EuroSAT} & gBase &32.79&62.70&71.25&\textbf{73.60}&61.15&\underline{72.77}&Base&55.80&\underline{90.25}&84.28&88.44&70.35&\textbf{91.33}\\
         & gNovel &\textbf{46.50}&23.45&\underline{33.95}&19.40&09.00&32.32&Novel&66.90&31.30&\underline{53.53}&49.49&33.90&\textbf{67.00}\\
         & gHM&38.46&34.13&\textbf{45.99}&30.71&15.69&\underline{44.76}&HM&60.85&46.48&65.47&\underline{63.47}&45.75&\textbf{77.30}\\
         \midrule
        \multirow{3}{4em}{DTD} & gBase &43.50&60.60&\underline{64.80}&62.30&42.60&\underline{62.73}&Base&53.70&\underline{75.12}&74.73&73.80&51.35&\textbf{75.97}\\
         & gNovel&\underline{41.29}&27.05&40.45&27.05&18.30&\textbf{44.27}&Novel&55.60&37.08&\underline{48.39}&46.38&29.85&\textbf{57.90}\\
         & gHM&42.37&37.40&\underline{49.81}&37.72&25.60&\textbf{51.91}&HM&54.63&49.65&\textbf{58.74}&56.96&37.75&\textbf{65.72}\\
    \midrule
    \multirow{3}{4em}{Oxford Pets} & gBase &85.90&84.70&85.75&\underline{85.95}&67.65&\textbf{87.00}&Base&91.20&90.15&\textbf{92.57}&\underline{92.36}&77.60&91.90\\
         & gNovel &85.59&85.25&\textbf{90.45}&87.10&65.65&\underline{89.27}&Novel&93.90&90.70&\textbf{94.61}&94.48&79.40&\underline{94.57}\\
         & gHM&85.74&84.97&\underline{88.04}&86.52&66.63&\textbf{88.12}&HM&92.53&90.42&\underline{93.58}&93.41&78.49&93.22\\
    \midrule
    \multirow{3}{4em}{Stanford Cars} & gBase &48.29&\underline{64.70}&62.25&64.30&17.35&\textbf{68.20}&Base&55.50&68.89&63.28&\textbf{71.79}&26.35&\underline{71.43}\\
         & gNovel &\textbf{64.09}&48.05&\underline{59.20}&53.45&21.65&57.60&Novel&66.50&57.13&\textbf{66.92}&59.36&25.50&\underline{64.77}\\
         & gHM&55.08&55.15&\underline{60.69}&58.38&19.26&\textbf{62.45}&HM&60.50&62.46&65.05&\underline{64.99}&25.92&\textbf{67.94}\\
    \midrule
    \multirow{3}{4em}{Flowers102} & gBase &62.59&\underline{89.40}&85.70&88.80&65.00&\textbf{92.52}&Base&69.70&\underline{95.22}&91.45&94.71&73.75&\textbf{96.40}\\
         & gNovel &\textbf{68.30}&50.70&\underline{63.85}&52.75&10.85&61.62&Novel&73.90&59.53&\textbf{71.75}&68.86&19.75&\underline{70.30}\\
         & gHM&65.32&64.70&\underline{73.18}&66.18&18.60&\textbf{73.97}&HM&71.74&73.26&\underline{80.41}&79.74&31.16&\textbf{81.31}\\
    \midrule
    \multirow{3}{4em}{Food101} & gBase &75.80&73.80&\textbf{78.30}&76.30&32.65&\underline{77.97}&Base&83.10&81.70&\underline{83.90}&83.77&37.85&83.57\\
         & gNovel &\textbf{78.90}&68.50&\underline{78.25}&72.90&17.60&76.60&Novel&84.50&78.13&\underline{85.23}&83.74&27.15&84.13\\
         & gHM&\underline{77.32}&71.05&\textbf{78.27}&74.56&22.87&77.28&HM&83.79&79.88&\underline{84.56}&83.75&31.62&83.85\\
    \midrule
    \multirow{3}{4em}{FGVC Aircraft} & gBase &12.69&\textbf{24.15}&20.20&21.60&8.65&\underline{23.17}&Base&18.80&28.39&24.91&\textbf{30.17}&14.20&\underline{28.97}\\
         & gNovel &22.10&14.75&\textbf{18.20}&14.25&6.95&\underline{17.45}&Novel&26.00&20.02&\textbf{25.69}&19.70&9.05&\underline{25.33}\\
         & gHM&16.12&18.31&\underline{19.15}&17.17&7.71&\textbf{19.91}&HM&21.82&23.48&\underline{25.29}&23.84&11.05&\textbf{27.03}\\
    \midrule
    \multirow{3}{4em}{SUN397} & gBase &56.70&66.65&67.05&\underline{67.15}&54.25&\textbf{70.40}&Base&66.40&76.33&75.33&\underline{76.90}&63.25&\textbf{78.20}\\
         & gNovel &60.50&53.30&\underline{61.80}&56.50&45.85&\textbf{62.20}&Novel&70.10&62.89&72.25&68.09&57.50&\textbf{73.27}\\
         & gHM&58.54&59.23&\underline{64.32}&61.37&49.70&\textbf{66.05}&HM&70.10&68.96&\underline{73.76}&72.23&60.24&\textbf{75.65}\\
    \midrule
    \multirow{3}{4em}{Caltech101} & gBase &88.59&91.35&\underline{91.65}&91.50&79.35&\textbf{92.13}&Base&91.00&95.20&95.35&\textbf{95.72}&84.80&\underline{95.67}\\
         & gNovel &81.69&82.15&\textbf{88.05}&86.30&58.65&\underline{87.50}&Novel&90.60&87.55&\textbf{91.92}&89.92&65.65&\underline{91.13}\\
         & gHM&85.00&86.51&\textbf{89.81}&88.82&67.45&\underline{89.76}&HM &90.80&91.21&\textbf{93.60}&92.73&74.01&\underline{93.34}\\
    \midrule
    \multirow{3}{4em}{ImageNet} & gBase &59.09&63.90&63.75&\underline{64.55}&41.40&\textbf{65.05}&Base&64.40&68.5&67.67&\underline{69.13}&47.00&\textbf{69.20}\\
         & gNovel&57.29&55.60&\textbf{58.75}&57.15&36.05&\underline{57.85}&Novel &60.10&58.76&\textbf{62.45}&57.39&39.35&\underline{61.40}\\
         & gHM&58.18&59.46&\underline{61.15}&60.63&38.54&\textbf{61.24}&HM&62.18&63.29&\underline{64.96}&62.72&42.84&\textbf{65.07}\\
    \bottomrule
    \end{tabular}
    }
    \vspace{3pt}
    \caption{GZS benchmark and Base-to-Novel Generalization benchmark using ResNet backbone. Metrics for the GZS benchmark, such as gBase, gNovel, and gHM, are employed in the left section of the table. Conversely, metrics like Base, Novel, and HM are utilized to assess the Base-to-Novel benchmark in the right section. On average, our method outperforms all the baselines.  We regret the mistake in Tab~\ref{generalizedresnet} of the main paper, where we incorrectly stated our method outperformed CLIP in the GZS benchmark. This error will be rectified in the revised version.}
    \label{gzsresnetappendix}
\end{table*}

\begin{table}
    \centering
    \scalebox{0.81}{
    \begin{tabular}{lccccccccccc>{\columncolor[gray]{0.8}}c}
    \toprule
         &Source&\multicolumn{11}{c}{Target} \\
    \cmidrule(lr){2-2}\cmidrule(lr){3-13}
         & \rotatebox{90}{ImageNet}&\rotatebox{90}{Caltech101}&\rotatebox{90}{OxfordPets}&\rotatebox{90}{StanfordCars}&\rotatebox{90}{Flowers102}&\rotatebox{90}{Food101}&\rotatebox{90}{Aircraft}&\rotatebox{90}{SUN397}&\rotatebox{90}{DTD}&\rotatebox{90}{EuroSAT}&\rotatebox{90}{UCF101}&\rotatebox{90}{Average}\\
    \midrule
    CoOp&71.51&93.70&89.14&64.51&68.71&85.30&18.47&64.15&41.92&46.39&66.55&63.88\\
    CoCoOp&71.02&94.43&90.14&65.32&71.88&86.06&22.94&67.36&45.73&45.37&68.21&65.74\\
    
    VPT&70.60&91.80&90.40&63.70&67.30&83.10&22.70&66.10&46.10&37.10&65.90&63.42\\
    MaPLe&70.72&93.53&90.49&65.57&72.23&86.20&24.74&67.01&46.49&48.06&68.69&66.30\\

    KgCoOp&69.94&94.08&90.13&65.63&71.21&86.48&23.85&67.47&45.80&41.98&68.33&65.49\\
    ProGrad&62.17&88.30&86.43&55.61&62.69&76.76&15.76&60.16&39.48&28.47&58.70&57.36\\
    PSRC&71.27&93.60&90.25&65.70&70.25&86.15&23.90&67.10&46.87&45.50&68.75&65.81\\
    CLIP-VDT&68.10&85.40&83.50&50.30&56.00&72.50&14.60&56.30&42.70&24.70&53.80&53.98\\
    KAPT&N/A&88.90&89.40&58.15&68.00&79.95&17.95&N/A&44.80&41.35&65.05&61.50\\
    \midrule
    \methodname{} (Ours)&71.40&94.53&90.14&64.58&71.31&86.23&24.47&68.09&48.61&49.10&71.52&\textbf{66.85}\\
    \bottomrule
    \end{tabular}
    }
    \vspace{3pt}
    \caption{Cross-Dataset Generalization benchmark. Models are trained on Imagenet and tested on the entire label space of new datasets without fine-tuning. \methodname{} outperforms all baselines on average. N/A: not available in~\citep{kapt}. }

    \label{crossdatasetappendix}
\end{table} 

\newpage
\section{Generation of Class Descriptions}
\label{app sec attribute priors}
Tab.~\ref{tab:sampleattributes} shows class names sampled from different datasets and their respective descriptions retrieved using GPT-3.5~\citep{gpt35}. We use the query -- \texttt{``What are useful visual features for distinguishing a [classname] in a photo? Answer concisely.''}
Class descriptions differ from well-curated attributes found in datasets with annotated attributes such as AwA~\citep{awa} and CUB~\citep{cub} in three ways: (i) Our class descriptions may be noisy since no manual curation is used; (ii) They may not necessarily contain class-discriminative information, especially for similar classes; and (iii) Descriptions of a class are generated independently, and may not contain comparative traits w.r.t. other classes. These choices are primarily to keep our approach low-cost while integrating these finer details into fine-tuning of VLMs. It's important to note that our description generation occurs at the class level, not the image level, making it cost-efficient.

\begin{table}
    \centering
    \scalebox{0.7}{
    \begin{tabular}{c|l|c|l}
    \toprule
       \textbf{Class}  & \textbf{Descriptions} &\textbf{Class}  & \textbf{Descriptions} \\
       \textbf{(Dataset)}&&\textbf{(Dataset)}\\
    \midrule
       Breast stroke & 1. Arms moving in a circular motion&Diving & 1. Person in mid-air or jumping\\
        (UCF101) & 2. Kicking legs in a frog-like motion&(UCF101)& 2. Person wearing diving gear\\
        & 3. Head above water during stroke && 3. water splashing or ripples \\
        & 4. Positioned horizontally in the water&& 4. Person wearing gogglesr\\
        & 5. Pushing water forward and outwards&& 5. Person wearing swim cap \\
    \midrule
     Highway or road & 1. Long and straight path&Permanent cropland &1.  Uniform vegetation or crops\\
       (EuroSAT)  & 2. Multiple lanes for traffic&(EuroSAT)& 2. Irrigation systems or canals  \\
        & 3. Traffic signs&& 3. Organized rows or patterns \\
        & 4. Smooth and paved surface&& 4. Fences or boundaries  \\
        & 5. Guardrails or barriers&& 5. Distinct crop types or varieties  \\
    
    \midrule
    Striped& 1. Alternating bands or lines&Wrinkled & 1. Irregular and uneven surface\\
      (DTD)   & 2. Regular pattern of stripes&(DTD)& 2. Creases or folds \\
        & 3. Varying widths of stripes&& 3. Shadows indicating unevenness \\
        & 4. Contrasting colors between stripes&& 4. Lack of smoothness  \\
        & 5. Horizontal, vertical, diagonal stripes&& 5. Distorted or crumpled appearance  \\
    \midrule
     Maine coon & 1. Large domestic cat&Chihuahua & 1. Small breed of dog\\
      (Oxford Pets)   & 2. Long, bushy tail&(Oxford Pets) & 2. Rounded apple-shaped head \\
        & 3. Tufted ears with lynx-like tips&& 3. Erect, pointy ears \\
        & 4. Rectangular body shape&& 4. Short snout  \\
        & 5. Tufted paws&& 5. Short legs and long tail  \\
\midrule
     2008 chrysler pt  & 1. Convertible top&2012 ferrari ff coupe & 1. Sleek and sporty design\\
      cruiser convertible   & 2. Chrome grille&(Stanford Cars) & 2. Large and stylish alloy wheels \\
       (Stanford Cars) & 3. PT cruiser badge&& 3. Low and wide stance \\
        &4. Alloy wheels&& 4. Ferrari logo on the front and rear  \\
        & 5. Boxy shape&& 5. Dual exhaust pipes  \\
\midrule
     Watercress & 1. Small, round-shaped leaves&Trumpet creeper & 1. Bright orange or red flowers\\
      (Flowers102)   & 2. Vibrant green color&(Flowers102)& 2. Trumpet-shaped blossoms \\
        & 3. Thin, delicate stems&& 3. Long, tubular petals \\
        & 4. Water or moist environments&& 4. Green leaves with serrated edges \\
        & 5. Clusters of small white flowers&& 5. Hummingbirds and bees  \\
\midrule
     Hot dog & 1. Cylindrical-shaped food&Sushi & 1. Bite-sized and compact\\
       (Food101)  & 2. Bun or bread&(Food101)& 2. Rice as a base  \\
        & 3. Sausage or frankfurter&& 3. Raw or cooked fish \\
        & 4. Visible grill marks&& 4. Seaweed wrapping (nori)  \\
        & 5. Toppings like onions or relish&& 5. Served with soy sauce  \\
\midrule
     737-200 & 1. Two engines on the wings&Industrial area& 1. Factories or warehouses\\
      (FGVC Aircraft)   & 2. Low wing configuration& (SUN397)& 2. Smokestacks or chimneys  \\
        & 3. Narrow body&& 3. Cranes or heavy machinery \\
        & 4. Distinctive short fuselage&& 4. Conveyor belts or assembly lines \\
        & 5. Swept-back wings&& 5. Trucks or shipping containers  \\
\midrule
     Gramophone & 1. Phonograph Cylinder or Disc&Buckle & 1. Metal or plastic object\\
      (Caltech101)   & 2. Horn Speaker&(Imagenet)& 2. Rectangular or circular shape  \\
        & 3. Hand-Cranked Operation&& 3. Fastening or securing \\
        & 4. Nostalgic and Vintage Appeal&&4. Opened and closed  \\
        & 5. Vinyl or Shellac Records&& 5. Found on belts or straps  \\
    \bottomrule
        \end{tabular}
        }
        \vspace{3pt}
    \caption{Sample classes from various datasets and the corresponding descriptions provided by GPT-3.5.}
    \label{tab:sampleattributes}
\end{table}

\section{Limitations and Broader Impact}
\label{sec:limitations}
A key dependency of our framework is the need for an LLM to provide descriptions at a class level. We however believe that this has become increasingly feasible in recent times, especially since we require at a class level and not at the image level. Our work deals with learning prompts for generalizable image classification by leveraging cheaply available semantic knowledge in the form of class descriptions. We believe that our work can serve as a stepping stone for incorporating semantic information to solve multi-modal tasks like captioning and VQA. To the best of our knowledge, there are no direct detrimental effects of our work.

\end{document}